\documentclass[10pt,twocolumn,letterpaper]{article}

\usepackage[pagenumbers]{cvpr} %
\makeatletter
\@namedef{ver@everyshi.sty}{}
\makeatother
\usepackage{graphicx}
\usepackage{amsmath}
\usepackage{breqn}
\usepackage{amssymb}
\usepackage{booktabs}
\usepackage{siunitx}
\usepackage{tabularx}
\usepackage{color, colortbl}
\usepackage{svg}
\usepackage{tikz}
\usepackage{multirow}
\usepackage{adjustbox}

\usepackage[pagebackref,breaklinks,colorlinks]{hyperref}

\usepackage[capitalize]{cleveref}
\crefname{section}{Sec.}{Secs.}
\Crefname{section}{Section}{Sections}
\Crefname{table}{Table}{Tables}
\crefname{table}{Tab.}{Tabs.}

\usepackage{letltxmacro}

\newcolumntype{H}{>{\setbox0=\hbox\bgroup}c<{\egroup}@{}}

\definecolor{purple}{RGB}{160, 32, 240}
\definecolor{washblue}{RGB}{186, 224, 228}
\definecolor{sky}{RGB}{128, 128, 128}
\definecolor{seagreen}{RGB}{60, 179, 113}
\definecolor{building}{RGB}{128, 0, 0}
\definecolor{road}{RGB}{128, 64, 128}
\definecolor{sidewalk}{RGB}{0, 0, 192}
\definecolor{fence}{RGB}{64, 64, 128}
\definecolor{vegetation}{RGB}{128, 128, 0}
\definecolor{car}{RGB}{64, 0, 128}
\definecolor{sign}{RGB}{192, 128, 128}
\definecolor{pedestrian}{RGB}{64, 64, 0}
\definecolor{cyclist}{RGB}{0, 128, 192}

\def\be {\begin{equation}}
\def\ee {\end{equation}}
\def\beas {\begin{eqnarray*}}
\def\eeas {\end{eqnarray*}}
\def\bea {\begin{eqnarray}}
\def\eea {\end{eqnarray}}
\def\bes {\begin{equation*}}
\def\ees {\end{equation*}}

\DeclareMathOperator*{\argmax}{argmax}

\makeatletter
\usepackage{xspace}
\def\@onedot{\ifx\@let@token.\else.\null\fi\xspace}
\DeclareRobustCommand\onedot{\futurelet\@let@token\@onedot}

\makeatother

\newcommand{\method}{PEANUT}

\begin{document}

\title{PEANUT: Predicting and Navigating to Unseen Targets}

\author{Albert J. Zhai, ~~~ Shenlong Wang\\
University of Illinois at Urbana-Champaign\\
{\tt\small \{azhai2, shenlong\}@illinois.edu}
}
\maketitle

\begin{abstract}
Efficient ObjectGoal navigation (ObjectNav) in novel environments requires an understanding of the spatial and semantic regularities in environment layouts. In this work, we present a straightforward method for learning these regularities by predicting the locations of unobserved objects from incomplete semantic maps. 
Our method differs from previous prediction-based navigation methods, such as frontier potential prediction or egocentric map completion, by directly predicting unseen targets while leveraging the global context from all previously explored areas.
Our prediction model is lightweight and can be trained in a supervised manner using a relatively small amount of passively collected data. Once trained, the model can be incorporated into a modular pipeline for ObjectNav without the need for any reinforcement learning. We validate the effectiveness of our method on the HM3D and MP3D ObjectNav datasets. We find that it achieves the state-of-the-art on both datasets, despite not using any additional data for training.

\end{abstract}

\section{Introduction}

Embodied visual navigation refers to a range of tasks that involve an agent navigating within the physical world based on visual sensory input \cite{deitke2022retrospectives}. One such task that has recently gained popularity is ObjectGoal navigation, also known as ObjectNav \cite{batra2020objectnav}. 
Here, the agent is placed in an unknown environment and is tasked with navigating to a specific object category (e.g. ``toilet''). 
ObjectNav is a fundamentally important task for embodied AI because it tests both scene understanding and long-term memory. It has immediate applications for helping people with disabilities find objects in their homes. It is also a necessary precursor for many other semantic tasks, such as instruction following \cite{anderson2018vision} and question answering \cite{das2018embodied}.

\begin{figure}[t]
\centering
\includegraphics[width=\linewidth, trim={7cm 6cm 6.3cm 4cm}, clip]{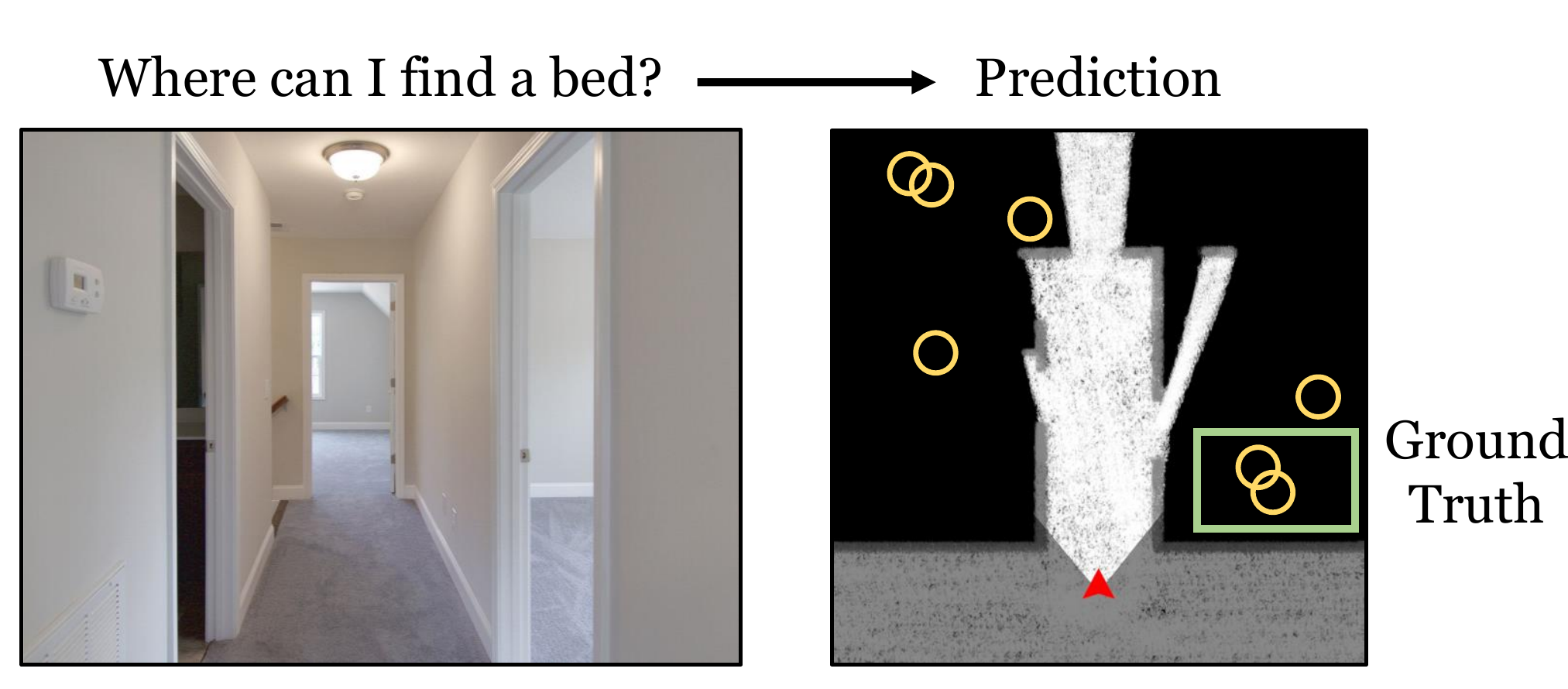}
\caption{\textbf{Explicit Prediction for ObjectNav.} 
To find a semantic target in a novel environment, an agent may make explicit guesses about where the target might be. Humans excel at this type of reasoning and can generate  a diverse set of reasonable guesses instantaneously. On the right is a top-down map representing the parts of the environment that have been observed. Yellow circles represent multiple human guesses as to where a bed might be.
}
\label{fig:teaser}
\vspace{-10pt}
\end{figure}

In ObjectNav, the agent is placed in a random location in an unknown environment and is given access to RGB-D observations, its own pose, and a target category. An episode is considered successful if the agent stops near an object of the target category within a given time limit. In order to perform this task efficiently, the agent must leverage priors about environment layouts while searching for the target. %
For example, imagine that you observe the image in Figure \ref{fig:teaser} and your target category is ``bed''. The room on the right does not have any visible furniture but has a good chance to be a bedroom. On the other hand, the room on the left has what appears to be a mirror above a countertop, suggesting that it is a bathroom that probably does not contain a bed. The room directly in front could be a bedroom, but it is farther away than the other two. Thus, it would be a good idea to look inside the room on the right first and the room directly in front second. An ideal ObjectNav agent should be able to make such predictions about unexplored areas and reason about the uncertainty in its predictions to plan efficient search routes.

\begin{figure*}[!t]
\centering
\includegraphics[width=\linewidth, trim={0.5cm 7cm 2.5cm 5cm}, clip]{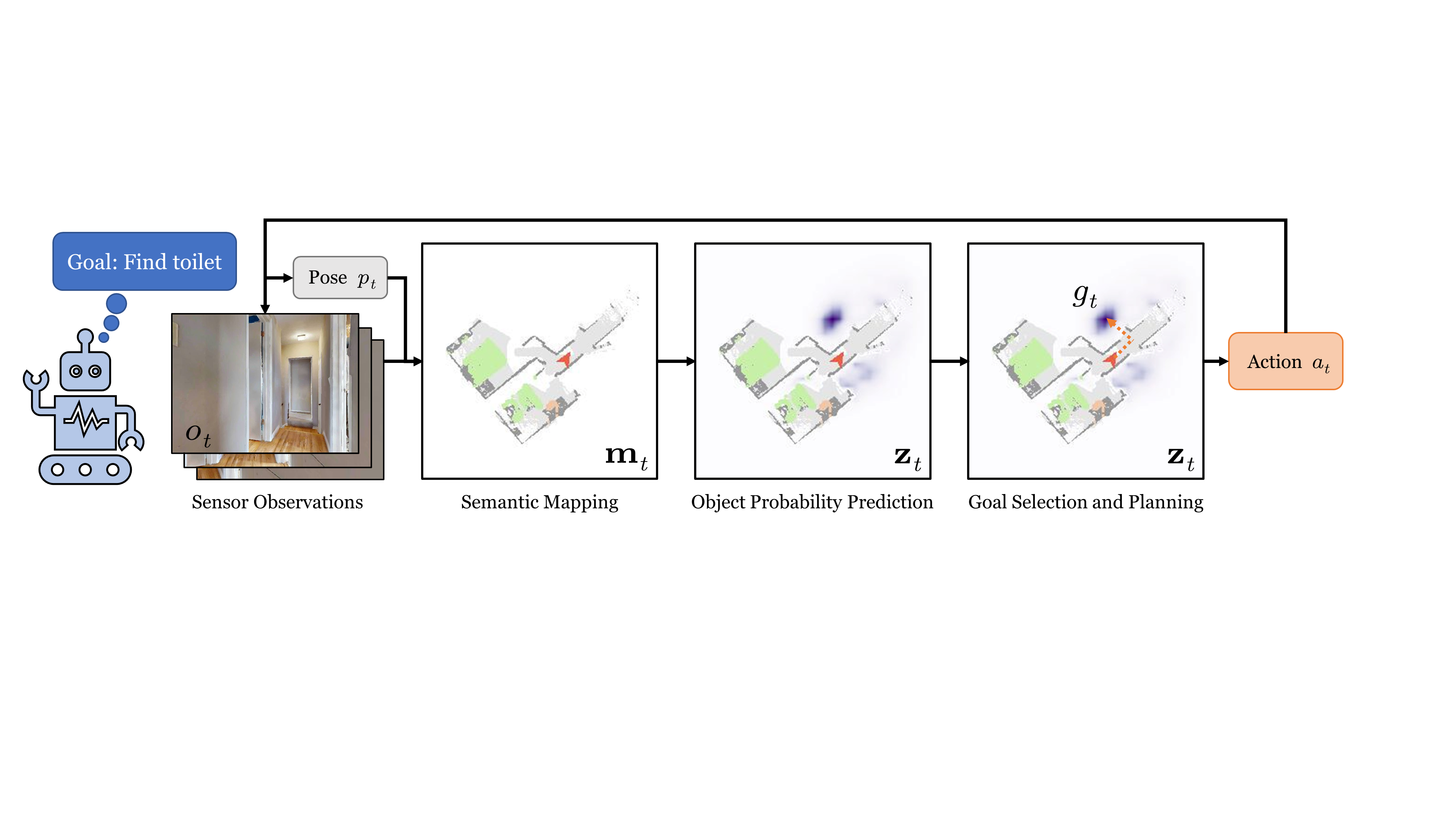}
\caption{\textbf{Overview of \method{}.} At each step, the agent's RGB-D observation $o_t$ and pose observations $p_t$ are used to update the incomplete global semantic map $\mathbf{m}_t$. This map is then used to predict a target object probability map $\mathbf{h}_t$, which is used to select long-term goals $g_t$. Finally, an analytical local planner is employed to calculate the low-level actions necessary to reach $g_t$. }
\label{fig:overview}
\end{figure*}

Existing methods for ObjectNav can be divided into two categories: end-to-end learning methods and modular methods. End-to-end learning methods aim to learn a policy that directly maps sensor observations to actions. The policy is usually modeled as a recurrent neural network and learned either using reinforcement learning \cite{wijmans2019dd, ye2021auxiliary, maksymets2021thda, khandelwal2022simple, deitke2022procthor} or imitation learning \cite{ramrakhya2022habitat}.
End-to-end learning methods are flexible in the policies that they can produce, but they require large amounts of data, incur high computational costs (especially for RL), and lack interpretability. Modular methods combat these issues by decomposing the task into subproblems. Current modular methods for ObjectNav start by building a semantic map of the environment \cite{cartillier2021semantic, chaplot2020object}, and then employ one module for high-level goal selection (``where to look'') and another module for low-level action planning (``how to get there''). 

The ``how to get there'' subproblem has been studied under the name of PointNav and is generally considered to be a solved problem \cite{deitke2022retrospectives}. The key question for ObjectNav is how to solve the ``where to look'' subproblem. Recently, several works have proposed to tackle the ``where to look'' subproblem as an explicit prediction task by training a network to make predictions about what lies in the unexplored areas of the environment using supervised learning \cite{liang2021sscnav, georgakis2021learning, ramakrishnan2022poni}. This explicit-prediction paradigm is especially attractive because it enables one to evade the sample inefficiency and high computational cost of RL \cite{georgakis2021learning, ramakrishnan2022poni}. 

Previous explicit-prediction methods for ObjectNav have tried to predict unseen targets from single-view observations \cite{georgakis2021learning, liang2021sscnav} or estimate frontier potential functions at the frontiers of the global semantic map \cite{ramakrishnan2022poni}. We believe using the context provided by the global semantic map is crucial for making informative predictions. However, we choose to directly predict unseen targets instead of frontier potential functions, since frontiers vary greatly depending on unseen obstacles and are thus difficult to predict accurately.

In this paper, we introduce PrEdicting And Navigating to Unseen Targets (\method{}), a novel explicit-prediction method for ObjectNav that predicts target object probabilities in the unexplored areas of the environment based on the agent's semantic map. Unlike previous map prediction methods \cite{liang2021sscnav, georgakis2021learning} that rely on single-view egocentric context, \method{} performs prediction in a global, allocentric context. This approach allows the agent to leverage information from previous timesteps and is consistent with recent cognitive models of human and animal navigation \cite{bush2015using, erdem2012goal}.  

In order to select long-term goals, we take the goal probability prediction and apply a simple distance-weighting scheme that encourages the agent to search closer locations before moving on to faraway ones. This produces a value map, %
 which we select long-term goals from by simply taking the argmax. Combining this goal selection module with an analytical low-level planner yields a modular pipeline for ObjectNav that does not require any reinforcement learning and can be trained in less than one GPU day.
 
 We evaluate \method{} on the Habitat-Matterport3D (HM3D) dataset \cite{ramakrishnan2021hm3d} and the Matterport3D (MP3D) dataset, which are the settings for the 2022 and 2021 Habitat ObjectNav Challenges respectively. On HM3D, \method{} outperforms all previous published methods, including methods that utilize additional datasets for training. On MP3D, \method{} also outperforms all previous methods, including recent modular methods. In our ablation study, we perform experiments to demonstrate the benefit of using global context for prediction.

\section{Related Work}

\paragraph{ObjectGoal Navigation.} ObjectNav remains to be a challenging task and is an active area of research in embodied AI. Existing methods for ObjectNav can be classified as either end-to-end learning methods or modular methods.  End-to-end learning methods seek to learn a mapping from sensor observations to low-level actions using a single neural network, usually consisting of a visual encoder followed by a recurrent cell that aggregates information over time. Most end-to-end learning methods train the network using reinforcement learning (RL), although recently, Habitat-Web \cite{ramrakhya2022habitat} collected a dataset of 80K human demonstrations and achieved strong performance using imitation learning. Significant progress has been made in end-to-end RL by introducing auxiliary tasks \cite{ye2021auxiliary}, different reward functions \cite{ye2021auxiliary, maksymets2021thda}, and different visual encoders such as CLIP-pretrained backbones \cite{khandelwal2022simple}, DINO-pretrained backbones \cite{yadav2022offline}, and graph convolutional networks on object graphs \cite{yang2018visual, du2020learning} . Many recent works focus on finding ways to leverage additional data for training \cite{deitke2022procthor, yadav2022offline, chang2020semantic, maksymets2021thda}. Currently, the state-of-the-art method for ObjectNav is ProcTHOR \cite{deitke2022procthor}, which generated thousands of synthetic scenes and applied end-to-end RL with a CLIP backbone. 

Modular methods seek to build a hierarchical policy in which one module handles high-level goal selection (``where to look'') and another module handles low-level action planning (``how to get there'') The advantages of this decomposition are that 1) the individual problems are easier to solve, 2) their solutions be reused across different tasks, and 3) the final pipeline is more interpretable than an end-to-end policy. In particular, the ``how to get there'' subproblem is equivalent to the task known as PointGoal Navigation (PointNav), for which near-perfect performance has been achieved by existing methods \cite{wijmans2019dd, kadian2020sim2real}. 

However, the ``where to look'' subproblem is unsolved and is the primary focus of research on modular ObjectNav. One approach is to treat this purely as an undirected exploration problem and try to maximize the area explored by the agent \cite{chaplot2019learning, luo2022stubborn, yamauchi1997frontier}.  This lends itself to simple heuristics that can perform well without any learning, but is ultimately limited in its efficiency. To leverage semantic priors for goal selection, SemExp \cite{chaplot2020object} built a top-down semantic map and used it to train an RL-based policy that minimizes distance to target objects, winning the 2020 Habitat ObjectNav Challenge. More recently, a few works \cite{ramakrishnan2022poni, georgakis2021learning} have proposed to select goals directly from supervised predictions about unexplored areas of the environment, avoiding RL entirely. This is the paradigm that we use in this paper. We discuss explicit prediction methods further in the next subsection.

\paragraph{Explicit Prediction for Navigation.} 
Efficient navigation in unknown environments requires prior models of environment layouts. End-to-end learning methods for navigation can make implicit predictions about unobserved areas of the environment, as suggested by visualizations of learned feature maps \cite{gupta2017cognitive}. Several works in recent years have aimed to predict specific properties of the environment layout for which explicit supervision can be obtained, allowing for straightforward offline training procedures. 
    
Such explicit-prediction methods have appeared in a variety of navigation tasks, including PointNav \cite{ramakrishnan2020occupancy}, RoomNav \cite{narasimhan2020seeing}, ImageNav \cite{chaplot2020neural, hahn2021no}, and ObjectNav \cite{liang2021sscnav, ramakrishnan2022poni, georgakis2021learning}. The prediction outputs range from occupancy maps \cite{ramakrishnan2020occupancy} to distances to the target \cite{ramakrishnan2022poni, hahn2021no} to semantic maps \cite{liang2021sscnav, georgakis2021learning}. The works that predict semantic maps for ObjectNav are the most similar to our approach. SSCNav \cite{liang2021sscnav} predicts the unexplored semantic map in a window around the agent along with a confidence map, and uses them as input to an RL-based goal selection policy. L2M \cite{georgakis2021learning} uses an ensemble of map predictors and selects goals by maximizing an upper confidence bound on the probability of the target category. However, both of these methods make predictions in an egocentric window using sensor inputs from only one timestep, whereas our method makes predictions on the allocentric global map using the information gathered from all previous timesteps. Another method, PONI \cite{ramakrishnan2022poni}, predicts potential functions (a combination of the connected free space and the proximity to the target) at the frontiers of the map in a global manner. We demonstrate in our experiments that directly predicting target object probabilities outperforms predicting frontier potentials in terms of navigation performance.

\section{Approach}

\paragraph{Problem Formulation.} ObjectNav is an episodic navigation task. In each episode, the agent starts at a random location in an unknown environment and is given a target category $c_{target} \in \{1, 2, \dots, C\}$, where $C$ is the number of possible target categories. At each timestep $t$, the agent observes an RGB-D image $o_t$ and a noiseless $(x, y, \theta)$ pose reading $p_t$, and then must output one of four discrete actions: \texttt{MOVE\_FORWARD}, \texttt{TURN\_LEFT}, \texttt{TURN\_RIGHT}, and \texttt{STOP}. An episode is considered successful if the agent executes $\texttt{STOP}$ within $\SI{1.0}{\metre}$ of a target object and the object can be viewed from the agent's position. Each episode has a time limit of $500$ steps.

\paragraph{Overview.} 
Our proposed method, \method{}, is a modular approach for ObjectNav (Fig.~\ref{fig:overview}). Like most modular approaches, \method{} uses a top-down semantic map as its internal representation of the environment. At the heart of our method is a novel unseen object prediction model. It addresses the question of ``where to look'' by making explicit predictions of the probability of the target appearing in each unexplored location of the semantic map, extracting a value map from these predictions, and then selecting the highest-value location as our goal. We train our prediction model via supervised learning using a dataset of actual semantic maps collected by a passive exploration agent.

We decompose the \method{} pipeline into four components: semantic mapping, unseen object probability prediction, prediction-based goal selection, and low-level planning. 
For semantic mapping, we use the same procedure as prior works \cite{chaplot2020object, ramakrishnan2022poni}, which we review in section \ref{mapping}. In section \ref{prediction}, we describe our explicit prediction model and how we gather data for training the model. In section \ref{goalselection}, we describe our simple strategy for selecting the agent's long term goal based on the prediction output. In section \ref{planning}, we describe the analytical local policy we use for planning actions to reach the goal.

\begin{table*}[]
    \centering
    \resizebox{\linewidth}{!}{
\setlength{\tabcolsep}{0.1em} %
\renewcommand{\arraystretch}{1.}
    \begin{tabular}{cccc}
          \tikz{
        \node[draw=black, line width=.5mm, inner sep=0pt] 
            {\includegraphics[width=.22\linewidth]{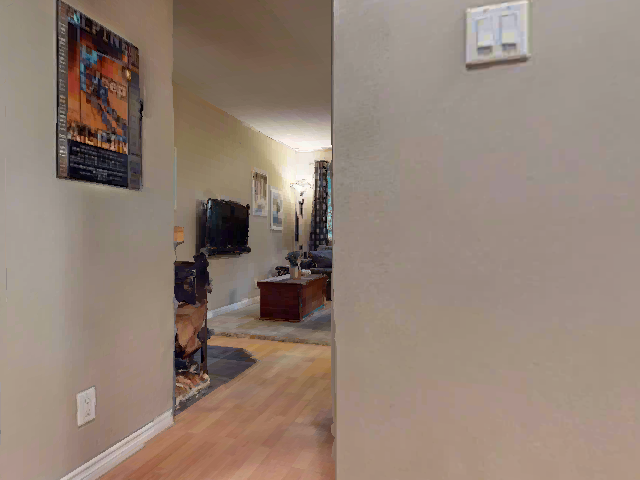}};
            \node[draw=black, draw opacity=1.0, line width=.3mm, fill opacity=0.8,fill=white, text opacity=1] at (-0.85 , 1.15) { \ Target: sofa \ };
        } &
        \tikz{
        \node[draw=black, line width=.5mm, inner sep=0pt] 
            {\includegraphics[width=.22\linewidth]{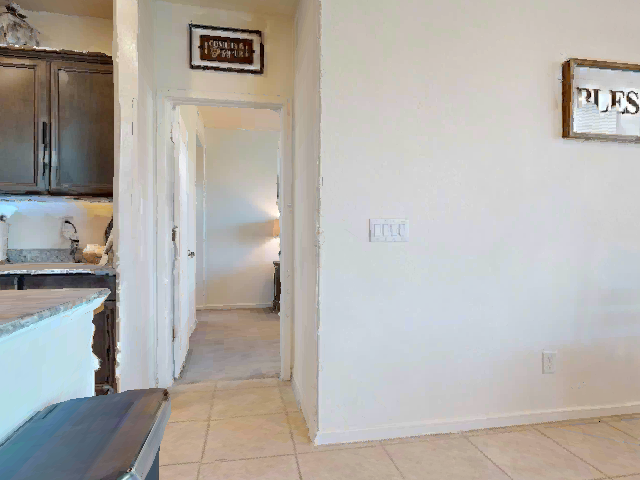}};
            \node[draw=black, draw opacity=1.0, line width=.3mm, fill opacity=0.8,fill=white, text opacity=1] at  (-0.9 , 1.15) { \ Target: bed \ };
        }& 
        \tikz{
        \node[draw=black, line width=.5mm, inner sep=0pt] 
            {\includegraphics[width=.22\linewidth]{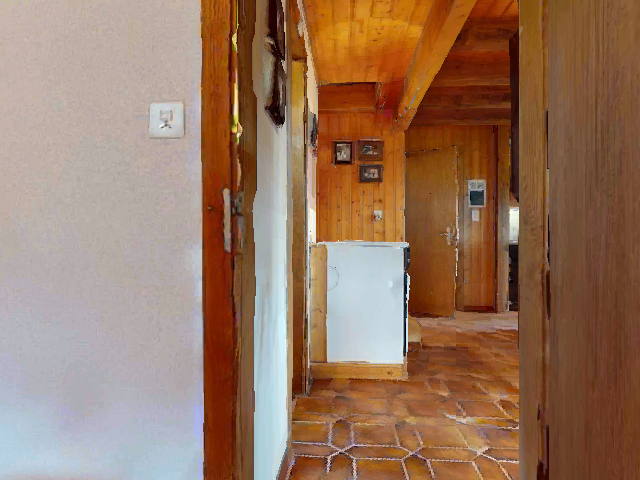}};
            \node[draw=black, draw opacity=1.0, line width=.3mm, fill opacity=0.8,fill=white, text opacity=1] at  (-0.8 , 1.15) { \ Target: chair \ };
        }& 
       \tikz{
        \node[draw=black, line width=.5mm, inner sep=0pt] 
            {\includegraphics[width=.22\linewidth]{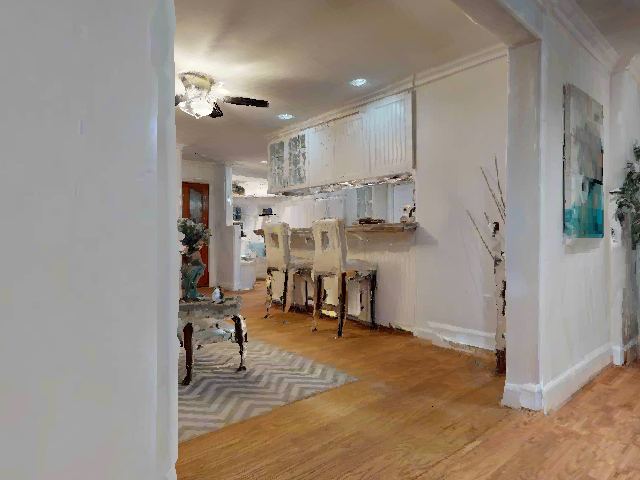}};
            \node[draw=black, draw opacity=1.0, line width=.3mm, fill opacity=0.8,fill=white, text opacity=1] at  (-0.85 , 1.15) { \ Target: sofa \ };
        } 
        \\
        
        \tikz{
        \node[draw=black, line width=.5mm, inner sep=0pt] 
        {\includegraphics[trim=1cm 0.25cm 0.5cm 1.25cm, clip,width=.22\linewidth]{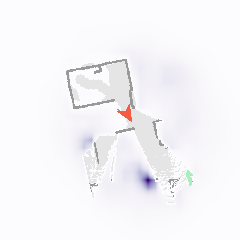}};
        } &
        \tikz{
        \node[draw=black, line width=.5mm, inner sep=0pt] 
        {\includegraphics[trim=1.5cm 0.6cm 0cm 0.9cm, clip,width=.22\linewidth]{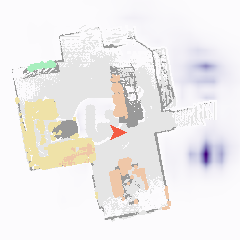}};
        } & 
        \tikz{
        \node[draw=black, line width=.5mm, inner sep=0pt] 
        {\includegraphics[trim=1.25cm 0.5cm 0.25cm 1cm, clip,width=.22\linewidth]{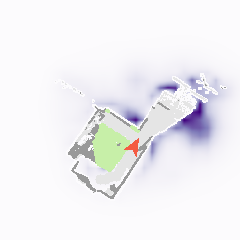}};
        } & 
        \tikz{
        \node[draw=black, line width=.5mm, inner sep=0pt] 
        {\includegraphics[trim=0.75cm 0.75cm 0.75cm 0.75cm, clip,width=.22\linewidth]{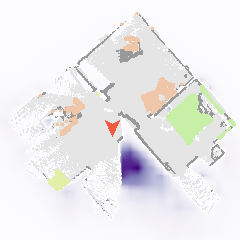}};
        } \\
        
    \end{tabular}
    }
\captionof{figure}{\textbf{Example target predictions}. We visualize some predictions made by our model in four different scenes from HM3D (val). The top row shows the agent's RGB observation, and the bottom row shows the incomplete semantic map overlaid with the target probability prediction. The model has learned to use semantic cues to improve its predictions, as demonstrated by its ability to predict the sofa facing the TV on the left. It is also able to produce multimodal predictions, which is especially apparent in the ``bed'' prediction. Note that the prediction network only takes the semantic map as input, not the RGB images.
}
\label{fig:example_preds}
\end{table*}

\subsection{Semantic Mapping} \label{mapping}
Our semantic mapping module follows the same procedure as previous works in semantic exploration \cite{chaplot2020object, ramakrishnan2022poni}. The module assumes access to RGB-D observations and the agent's pose. At each timestep, the RGB-D image is segmented into semantic categories using an existing segmentation model. The set of categories must include at least the $C$ target categories, but may also include other categories. Then, the depth observation is used to lift each pixel along with its semantic label to 3D. Points that are within a certain height range (based on the agent's height) are marked as obstacles. The point cloud is then converted into a voxel occupancy grid and is summed across the height dimension to obtain an egocentric map. This egocentric map is transformed into an allocentric coordinate system using the agent's pose and is aggregated with the existing global map.

In total, the semantic map $\mathbf{m}_t$ has $(N + 4) \times H \times W$ elements, where $N$ is the number of semantic categories. Channels 1 and 2 represent the obstacle map and the explored areas. Channels 3 and 4 contain the agent's current location and all past locations. The next $C$ channels correspond to the $C$ target semantic categories. The other $N - C$ channels represent the remaining semantic categories. 

\subsection{Object Probability Prediction} \label{prediction}
The core component of our navigation pipeline is the explicit prediction model, which represents the agent's belief of what lies outside the regions of the environment that it has explored. Most of the time, there are many possibilities as to what the environment layout may be, so the model must be able to express uncertainty in its predictions. Thus, we represent the prediction as another top-down map, aligned with $\mathbf{m}_t$, in which each element contains the probability of each target category appearing in that location. We train a network to predict these probabilities using the entire global map $\mathbf{m}_t$ as input, unlike previous works that predict based on observations from a single frame \cite{liang2021sscnav, georgakis2021learning}. We choose to directly predict target locations instead of potential functions \cite{ramakrishnan2022poni}, which vary greatly depending on the layout of unseen obstacles.  We hypothesize that by decoupling the prediction of goals from the prediction of obstacles, we are able to learn a useful prediction model more effectively from limited amounts of data.

\paragraph{Problem Formulation.} 
Given an incomplete semantic map and a set of target semantic categories, we wish to predict for every unexplored map location the probability of each target category appearing in that location.
Formally, let $\mathbf{m}_t \in \mathbb{R}^{(N + 4) \times H \times W}$ be the incomplete semantic map as described above and let $\mathcal{C} = \{1, 2, \dots, C\}$ be the set of possible target categories. 
Let $\mathbf{e}_t$ be an exploration mask indicating which locations have been explored, and let $\mathbf{M} \in \mathbb{R}^{C \times H \times W}$ be the ground-truth full semantic map of the environment containing the $C$ target categories. For each location $(i, j)$ satisfying $\mathbf{e}_t[i, j] = 0$ and each target category $c \in \mathcal{C}$, we wish to learn a model $f_\theta(c, i, j | \mathbf{m}_t) $ of the probability that an object of category $c$ exists at location $(i, j)$ in the complete map $\mathbf{M}$, given the information from the incomplete map $\mathbf{m}_t$.

\paragraph{Network Architecture and Loss Function.} We formulate this as a dense pixel prediction task and train the network to predict zero in the already-explored areas. This allows us to use any image-to-image network architecture for $f_\theta$.
In this work, we use a PSPNet \cite{zhao2017pyramid} architecture, but modify the number of input channels to $N+4$ and the number of output channels to $C$. A detailed description of the architecture is provided in the supplementary material. 

To train the network, we calculate training outputs $\mathbf{y}_t$ from ground-truth maps $\mathbf{M}$ by setting the already-explored areas to zero:
\begin{equation}\label{eq:1}
    \mathbf{y}_t[c, i, j] = (1 - \mathbf{e}_t[i, j]) \mathbf{M}[c, i, j].
\end{equation}
We then train using binary cross-entropy loss, averaged over the $C$ categories:
\begin{dmath}\label{eq:2} 
   L = \frac{1}{CHW} \sum_{c, i, j}  (\mathbf{y}_t[c, i, j] \log f_\theta(c, i, j | \mathbf{m}_t) +  (1 - \mathbf{y}_t[c, i, j]) \log (1 - f_\theta(c, i, j | \mathbf{m}_t) )).
\end{dmath}

\paragraph{Training Data Generation.} We generate training data for our prediction task from semantic maps collected during episodes of agent interaction. Using the mapping procedure described in section \ref{mapping}, we let an exploration agent wander around for 500 steps in different scenes from different starting locations and saved the incomplete semantic map every 25 steps. Here the agent follows the baseline exploration strategy described in \cite{luo2022stubborn} due to its empirical efficiency, but any reasonable exploration strategy can be applied instead. We use the maps from steps 25 to 250 as incomplete input maps $\mathbf{m}_t$ and generate $\mathbf{M}$ from the map at step 500, where the agent has usually explored the entire scene.

\paragraph{Inference.} During navigation, we simply apply the prediction network on the collected semantic map $\mathbf{m}_t$ and extract the probability map $\mathbf{z}_t$ corresponding to $c_{target}$:
\begin{equation}\label{eq:3}
    \mathbf{z}_t[i, j] = f_\theta(c_{target} , i, j | \mathbf{m}_t),
\end{equation}
and use it to select long-term goals for the agent. We update the prediction every 10 steps or whenever the previous goal has been reached.

\subsection{Prediction-Based Goal Selection} \label{goalselection} We provide a simple method for selecting long-term goals using the target probability predictions $\mathbf{z}_t$.
Perhaps the simplest way to plan using $\mathbf{z}_t$ is to set $g_t$ to be the location with the highest probability $\argmax_{(i, j)} ~ \mathbf{z}_t[i, j]$.  
However, it is usually more efficient for the agent to search in locations closer to its current location before moving on to locations farther away. Thus, for each location $(i, j)$, we use the Fast Marching Method \cite{fmm} to calculate the geodesic distance $d_t(i, j)$ from the agent's current location using the current obstacle map (channel 1 of $\mathbf{m}_t$). The geodesic distance to $(i, j)$ is the length of the shortest path to $(i, j)$, which is a better estimate of the time needed to reach $(i, j)$ compared to the Euclidean distance. We then use $d_t$ as an exponential weight factor for $\mathbf{z}_t$ and set $g_t$ to be the argmax of the weighted result. Formally,
\begin{equation}\label{eq:4}
    g_t = \argmax_{(i, j)} ~ \exp{(-d_t(i, j) / \lambda)}\mathbf{z}_t[i, j],
\end{equation}
where $\lambda$ is a tunable parameter that allows for a tradeoff between higher-probability locations and closer locations. 
Note that the calculated geodesic distance is not aware of any obstacles outside what the agent has currently seen -- it is fully
optimistic about traversability in unexplored areas. 

\subsection{Analytical Local Policy} \label{planning}
Our local policy converts the long-term goal $g_t$ into low-level actions using the same technique as previous works \cite{chaplot2019learning, chaplot2020object, ramakrishnan2022poni}. It computes the shortest path in terms of $(i, j)$ locations using the Fast Marching Method and then extracts a waypoint from this path using the agent's step distance. Replanning is done at every timestep. We also deploy the recovery behavior described in \cite{luo2022stubborn} when the agent gets stuck between obstacles.

\begin{table*}[]
    \centering
    \resizebox{\linewidth}{!}{
\setlength{\tabcolsep}{0.1em} %

    \begin{tabular}{ccccc}
        \tikz{
        \node[draw=black, line width=.5mm, inner sep=0pt] 
            {\includegraphics[width=.22\linewidth]{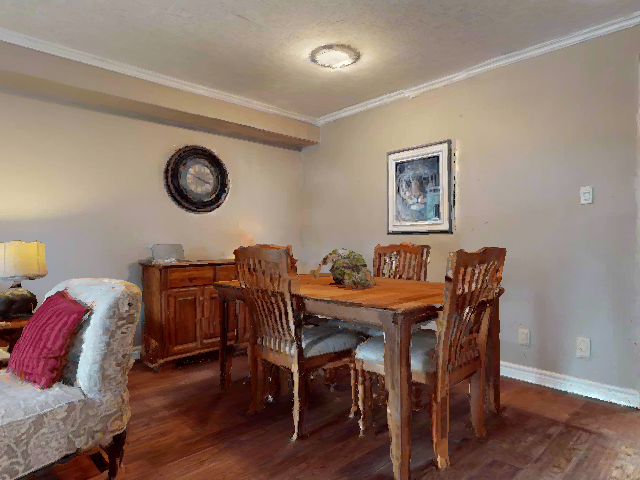}};
            \node[draw=black, draw opacity=1.0, line width=.3mm, fill opacity=0.8,fill=white, text opacity=1] at (-0.80 , 1.15) { \ Target: toilet \ };
        } &
        \tikz{
        \node[draw=black, line width=.5mm, inner sep=0pt] 
        {\includegraphics[width=.22\linewidth]{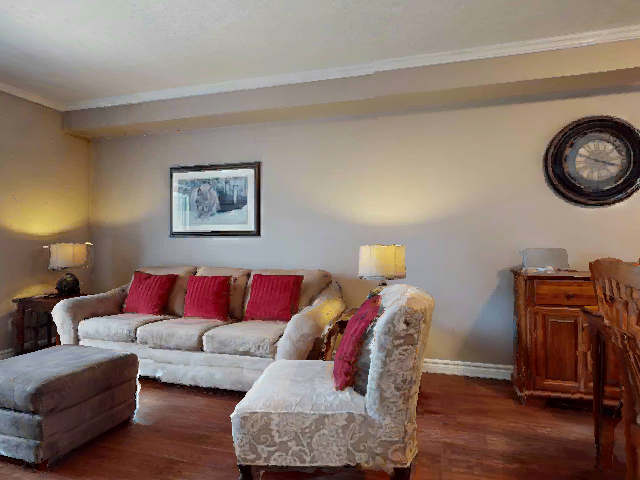}};
        } & 
        \tikz{
        \node[draw=black, line width=.5mm, inner sep=0pt] 
        {\includegraphics[width=.22\linewidth]{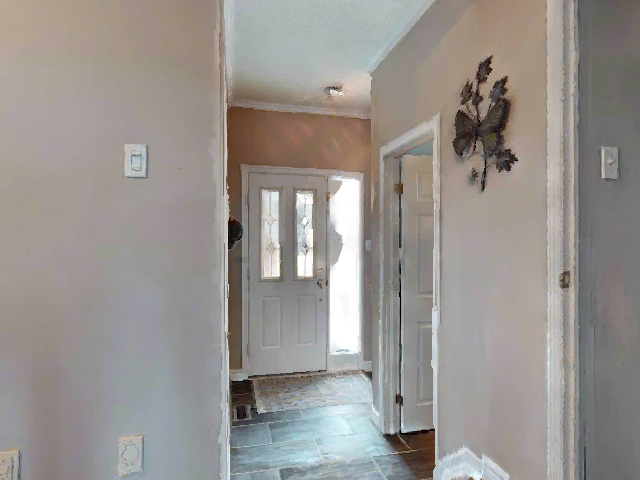}};
        } & 
        \tikz{
        \node[draw=black, line width=.5mm, inner sep=0pt] 
        {\includegraphics[width=.22\linewidth]{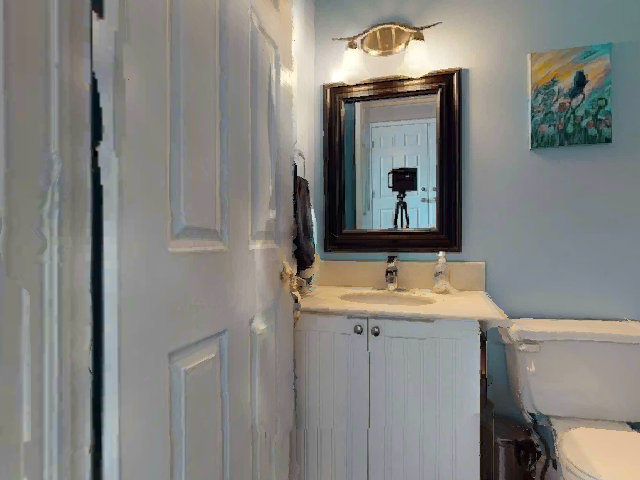}};
        } & 
        \tikz{
        \node[draw=black, line width=.5mm, inner sep=0pt] 
        {\includegraphics[width=.22\linewidth]{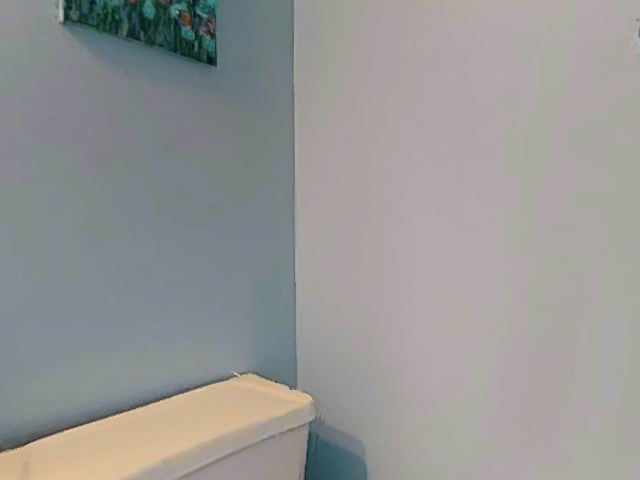}};
        } \\
        
        \tikz{
        \node[draw=black, line width=.5mm, inner sep=0pt] 
        {\includegraphics[trim=1.5cm 0cm 0cm 1.5cm, clip,width=.22\linewidth]{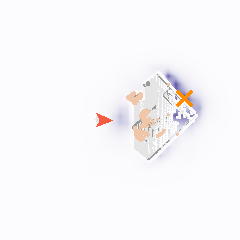}};
        } &
        \tikz{
        \node[draw=black, line width=.5mm, inner sep=0pt] 
        {\includegraphics[trim=1.5cm 0cm 0cm 1.5cm, clip,width=.22\linewidth]{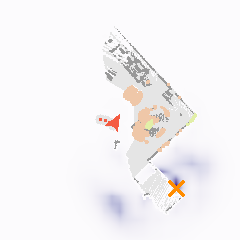}};
        } & 
        \tikz{
        \node[draw=black, line width=.5mm, inner sep=0pt] 
        {\includegraphics[trim=1.5cm 0cm 0cm 1.5cm, clip,width=.22\linewidth]{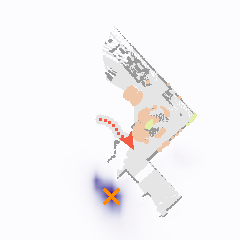}};
        } & 
        \tikz{
        \node[draw=black, line width=.5mm, inner sep=0pt] 
        {\includegraphics[trim=1.5cm 0cm 0cm 1.5cm, clip,width=.22\linewidth]{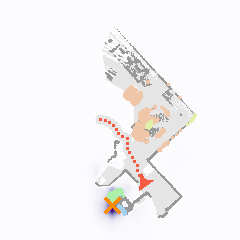}};
        } & 
        \tikz{
        \node[draw=black, line width=.5mm, inner sep=0pt] 
        {\includegraphics[trim=1.5cm 0cm 0cm 1.5cm, clip,width=.22\linewidth]{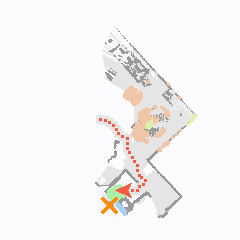}};
        } \\
        $t=1$ & 
        $t=10$ & 
        $t=20$ & 
        $t=49$ & 
        $t=61$
    \end{tabular}
    }
\captionof{figure}{\textbf{Example navigation episode using \method{}.} We visualize a full episode of navigation in a scene from HM3D (val). The top row shows the agent's RGB observation, and the bottom row shows the incomplete semantic map overlaid with the target probability prediction. The agent's selected long-term goal is marked by an orange cross. At $t = 1$, without having much information, the agent's prediction is somewhat random. At $t = 10$, it updates its prediction to assign higher probabilities around the hallway to the right, and begins to move over there. At $t = 20$, after seeing more of the hallway, it updates its prediction to sharply favor the open room on the right (which is indeed a bathroom).  At $t = 49$, the agent detects the toilet, and at $t = 61$, it ends the episode successfully.
}
\label{fig:example_episode}
\end{table*}

\section{Experiments}

We evaluate \method{}  on standard ObjectNav datasets~\cite{ramakrishnan2021hm3d, chang2017matterport3d} and show that it outperforms previous methods, including state-of-the-art methods that rely on additional data for training. We also perform several ablation experiments to study the effect of various design choices in our pipeline.

\subsection{Experimental Setup}

\paragraph{Datasets.} 
We conduct experiments on both the Habitat-Matterport3D (HM3D) dataset \cite{ramakrishnan2021hm3d} and the Matterport3D (MP3D) dataset  \cite{chang2017matterport3d} using the Habitat simulator \cite{habitat19iccv}. Both datasets contain 3D reconstructions of real-world indoor environments from which realistic RGB-D observations can be rendered. HM3D is more recent and is of slightly higher quality, but there are not as many publicly available results for it compared to MP3D. 

For HM3D, our setup is consistent with the 2022 Habitat ObjectNav Challenge \cite{habitatchallenge2022}, which has 6 goal categories and 80 train / 20 val / 20 test scenes. For MP3D, our setup is consistent with the 2021 Habitat ObjectNav Challenge \cite{batra2020objectnav}, which has 21 goal categories and 56 train / 11 val / 18 test scenes.  
In both settings, the agent's RGB-D observation has a resolution of $640 \times 480$ with a horizontal field of view of 79 degrees. The step distance for \texttt{MOVE\_FORWARD} is \SI{0.25}{\metre} and the turn angle is $30$ degrees.

\paragraph{Metrics.} 
We evaluate navigation performance using two standard metrics. \textbf{Success} is the proportion of episodes in which the agent successfully stopped at a goal object. \textbf{SPL} is the success weighted by the length of the agent’s path relative to the oracle shortest path length \cite{batra2020objectnav}.

\paragraph{Baselines.} 
For HM3D, we compare \method{} with four public methods. To the best of our knowledge, these are the only methods for which papers are available.

\begin{itemize}
    \item \textbf{DD-PPO} \cite{wijmans2019dd} applies end-to-end RL with large-scale distributed training.
    \item \textbf{Habitat-Web} \cite{ramrakhya2022habitat} applies end-to-end imitation learning using human teleoperated demonstrations.
    \item \textbf{OVRL} \cite{yadav2022offline} pretrains a visual encoder using DINO \cite{caron2021emerging} on images from the Omnidata dataset \cite{eftekhar2021omnidata} before applying end-to-end RL.
    \item \textbf{ProcTHOR} \cite{deitke2022procthor} applies end-to-end RL on a large procedurally generated dataset of synthetic scenes. It currently holds the state-of-the-art SPL on HM3D.
\end{itemize}

\noindent
For MP3D, we compare \method{} with eight other methods. Of these eight methods, five use end-to-end learning:

\begin{itemize}
    \item \textbf{DD-PPO} \cite{wijmans2019dd}, \textbf{Habitat-Web} \cite{ramrakhya2022habitat}, and \textbf{OVRL} \cite{yadav2022offline} are described above.
    \item \textbf{Red Rabbit} \cite{ye2021auxiliary} incorporates 6 auxiliary tasks for end-to-end RL. It is the winner of the 2021 Habitat ObjectNav Challenge.
    \item \textbf{TreasureHunt} \cite{maksymets2021thda} artificially inserts objects from the YCB dataset \cite{calli2017ycb} into MP3D scenes to provide additional data for end-to-end RL. 
\end{itemize}

\noindent
The other three methods are modular methods:
\begin{itemize}
    \item \textbf{ANS} \cite{chaplot2019learning} is a modular method that trains an RL-based goal selection policy to maximize area explored. 
    \item \textbf{PONI} \cite{ramakrishnan2022poni} is a modular method that predicts potential functions on the frontiers of a top-down map, and then selects goals by picking the highest-potential frontier. The potential function is a sum of the connected free space and the inverse distance to the target. 
    \item \textbf{Stubborn} \cite{luo2022stubborn} is a modular method that simply sets the goal to be one of four corners of a local crop of the map, rotating when reaching a dead-end. It also fuses target detections across frames. It is currently the best-performing modular method on MP3D.
\end{itemize}

\paragraph{Implementation Details.} 
Our semantic mapping module requires a 2D semantic segmentation model. For HM3D, we finetune a COCO-pretrained Mask-RCNN \cite{he2017mask} (on images from HM3D) to predict the 6 goal categories along with 3 other categories (\textit{fireplace}, \textit{mirror}, and \textit{bathtub}). For MP3D, we use the publicly available RedNet \cite{jiang2018rednet} model from \cite{ye2021auxiliary}, which predicts 21 object categories.

To train \method{}, we collect maps as described in section \ref{prediction} via Habitat. For fair comparisons with other works, we only train on scenes from MP3D when evaluating ObjectNav performance on MP3D, and likewise for HM3D, although better performance may be achieved by combining the two or adding other datasets. For both MP3D and HM3D, we collect sequences of maps from 50 starting locations for each scene. This gives \num{4000} train / \num{1000} val sequences for HM3D and \num{2800} train / \num{550} val sequences for MP3D. During training, we apply random rotation, flip, and padded-crop operations for data augmentation.

Our prediction network uses a PSPNet \cite{zhao2017pyramid} architecture with a ResNet50 \cite{he2016deep} backbone and an auxiliary loss weight of 0.4.
For both datasets, we use an Adam optimizer \cite{kingma2014adam} with $\alpha = 0.0005$, $(\beta_1, \beta_2) = (0.9, 0.999)$, and a batch size of 8, with $\alpha$ decaying according to the ``poly" learning rate policy described in \cite{zhao2017pyramid}. For HM3D, we stop training at \num{28000} iterations. For MP3D, we stop training at \num{22000} iterations. For goal selection, we set $\lambda = \SI{5}{\metre}$.

\begin{table}[!t]
\caption{ObjectNav results on HM3D (test-standard). The ``Ext. Data'' column indicates whether the method uses non-HM3D data when training its navigation policy.}
\small
\centering
\begin{tabular}{lccc}
\toprule
Method  & SPL ($\uparrow$) & Success ($\uparrow$) & Ext. Data  \\
\midrule
DD-PPO \cite{wijmans2019dd} & 0.12 & 0.26 & no \\
Habitat-Web \cite{ramrakhya2022habitat}  & 0.22 & 0.55 & yes \\
OVRL \cite{yadav2022offline} & 0.27 &	0.60 & no\\
ProcTHOR \cite{deitke2022procthor} & 0.32  & 0.54 & yes\\

\midrule
\method{} (Ours) & \textbf{0.33} & \textbf{0.64} & no\\
\bottomrule
\label{tab:hm3dtest}
\end{tabular}
\end{table}

\subsection{ObjectNav on HM3D}

\paragraph{Quantitative Results.}
We report results from the test-standard split of the 2022 Habitat ObjectNav Challenge leaderboard in Tab. \ref{tab:hm3dtest}. \method{} achieves an SPL of 0.33 and a success rate of 0.64, outperforming all other published methods in both metrics. Note that two of the other methods rely on additional data for training their navigation policy, in the form of procedurally generated scenes for ProcTHOR \cite{deitke2022procthor} and human teleoperated demonstrations for Habitat-Web \cite{ramrakhya2022habitat}. \method{} does not use any data outside of HM3D other than through the COCO-pretrained Mask-RCNN. By making globally-informed predictions about where the target objects may be, \method{} can select goals that help the agent search through the scene efficiently. 

\paragraph{Qualitative Results.}
We show example target predictions made by our prediction model in Fig. \ref{fig:example_preds}. We can see that the model has learned strong priors on the spatial regularities of indoor layouts, enabling it to make rather sharp predictions. In addition, it is able to exploit semantic regularities such as the co-occurrence of TVs and sofas. We can also confirm that the model is able to produce multimodal predictions, reflecting the inherent uncertainty in the prediction task.

We visualize a full episode of navigation using \method{} in Fig. \ref{fig:example_episode}. We observe that the prediction-based goal selection policy allows the agent to search for the target in an intelligent manner. We also notice that the prediction can change quite dramatically over the span of a few timesteps, suggesting that the model takes advantage of new information efficiently.

\begin{table}[!t]
\caption{ObjectNav results on MP3D (val). The ``Ext. Data'' column indicates whether the method uses non-MP3D data when training its navigation policy.}
\small
\centering
\begin{tabular}{lccc}
\toprule
Method  & SPL ($\uparrow$)   & Success ($\uparrow$) & Ext. Data \\
\midrule
DD-PPO \cite{wijmans2019dd} & 0.018 & 0.080 & no \\
Habitat-Web \cite{ramrakhya2022habitat} & 0.102 & 0.354& yes \\
OVRL \cite{yadav2022offline} & 0.074 &	0.286 & no\\
Red-Rabbit \cite{ye2021auxiliary} & 0.079  & 0.346 & no \\
TreasureHunt  \cite{maksymets2021thda} & 0.110 & 0.284 & yes \\

\midrule
ANS  \cite{chaplot2019learning} & 0.092 &  0.273 & no \\
PONI \cite{ramakrishnan2022poni}&  0.121 &  0.227 & no \\
Stubborn \cite{luo2022stubborn} & 0.149 & \textbf{0.407} & no \\
\midrule
\method{} (Ours) & \textbf{0.158} & 0.405 & no \\
\bottomrule
\label{tab:mp3dtest}
\end{tabular}
\end{table}

\subsection{ObjectNav on MP3D}
\paragraph{Quantitative Results.}
We report results on the MP3D val split (2195 episodes) of in Tab. \ref{tab:mp3dtest}. For DD-PPO \cite{wijmans2019dd} and ANS \cite{chaplot2019learning}, we report results from \cite{ramakrishnan2022poni}. For Stubborn \cite{luo2022stubborn}, we obtain results by running the official public code. For all other methods, we report results from their respective papers.
As with HM3D, \method{} outperforms all previous methods on MP3D in terms of SPL. Its success rate of 0.405 is matched only by Stubborn \cite{luo2022stubborn}, which fuses detection scores over time for more robust target detection.
\method{} significantly outperforms PONI \cite{ramakrishnan2022poni}, another explicit-prediction method. This is because frontier potentials vary widely depending on the layout of unexplored obstacles and are thus difficult to estimate accurately.

\paragraph{Qualitative Results}
We show example target predictions from \method{} and compare them with potential functions estimated by PONI \cite{ramakrishnan2022poni} in Fig. \ref{fig:example_poni}. The PONI predictions were obtained using their publicly available pretrained model. We find that \method{}'s predictions appear to be more informative and useful than PONI's, as PONI often predicts high potentials for frontiers that are not worth exploring (such as in the corners of already-explored rooms).

\begin{table}
    \centering{
\setlength{\tabcolsep}{0.1em} %
\renewcommand{\arraystretch}{1.}
    \begin{tabular}{ccc}
    \tikz{
        \node[draw=black, line width=.5mm, inner sep=0pt] 
            {\includegraphics[height=0.28\linewidth]{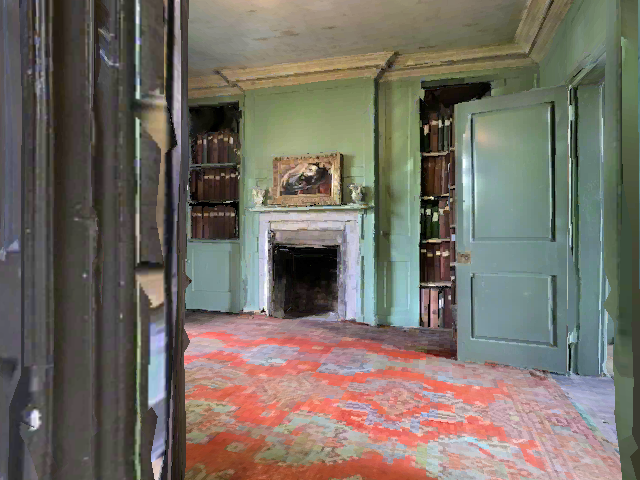}};
            \node[draw=black, draw opacity=1.0, line width=.3mm, fill opacity=0.8,fill=white, text opacity=1] at (-0.52, 0.88) { \ \small{Target: table} \ };
        } 
        & 
        \tikz{
        \node[draw=black, line width=.5mm, inner sep=0pt] 
        {\includegraphics[ clip,width=0.28\linewidth]{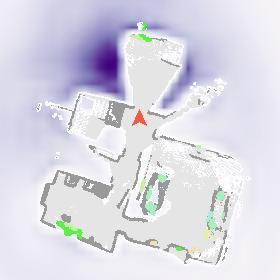}};
        } 
         &
        \tikz{
        \node[draw=black, line width=.5mm, inner sep=0pt] 
        {\includegraphics[ clip,width=0.28\linewidth]{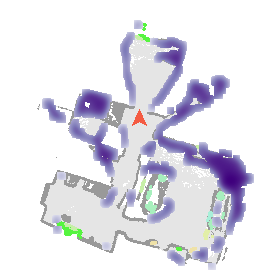}};
        } \\

        \tikz{
        \node[draw=black, line width=.5mm, inner sep=0pt] 
            {\includegraphics[height=0.28\linewidth]{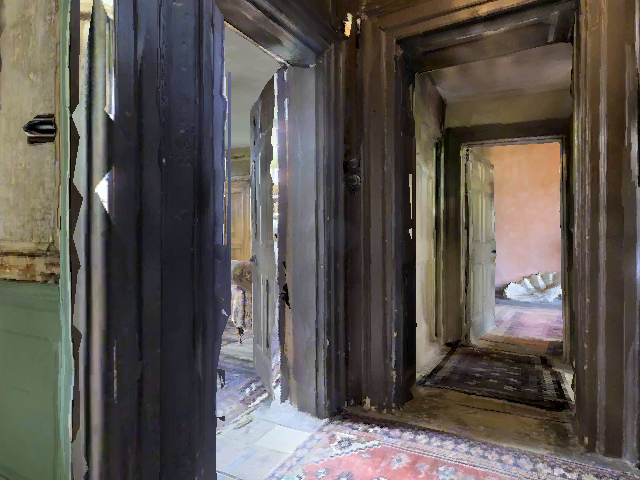}};
            \node[draw=black, draw opacity=1.0, line width=.3mm, fill opacity=0.8,fill=white, text opacity=1] at (-0.35, 0.88) { \ \small{Target: cushion} \ };
        } 
        & 
        \tikz{
        \node[draw=black, line width=.5mm, inner sep=0pt] 
        {\includegraphics[ clip,width=0.28\linewidth]{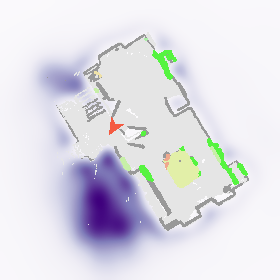}};
        } 
         &
        \tikz{
        \node[draw=black, line width=.5mm, inner sep=0pt] 
        {\includegraphics[ clip,width=0.28\linewidth]{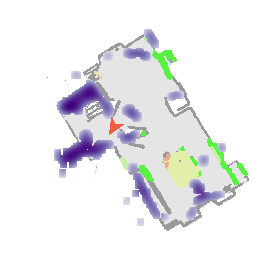}};
        } 
    \end{tabular}
    }
\captionof{figure}{\textbf{Qualitative comparison with potential functions (PONI) \cite{ramakrishnan2022poni}.} We visualize predictions made by \method{} on MP3D (val) and compare them with potential functions estimated by PONI. We observe that \method{} tends to make cleaner, sharper predictions than PONI, which often assigns high potentials to frontiers that clearly cannot lead to any target object.
}
\label{fig:example_poni}
\end{table}

\subsection{Ablation Studies}

\paragraph{Global Context for Prediction.}  We study the effect of leveraging full global context for prediction as opposed to only using single-frame observations for prediction, as done by SSCNav \cite{liang2021sscnav} and L2M \cite{georgakis2021learning}. We emulate egocentric single-frame maps by taking $\SI{6}{\metre} \times \SI{6}{\metre}$ crops of $\mathbf{m}_t$ from random locations along the agent's trajectory, and masking out everything outside of the agent's viewing frustum when projected onto the map. Note that this actually gives more information than an actual single-frame observation, since it does not account for occlusion from the agent's view. We apply random rotation and flipping and train a prediction network to predict object probabilities in the unexplored areas of the crop in the exact same manner as with the global context version. 

We compare prediction quality using two metrics: \textbf{DTO} looks at the highest-probability location from the prediction and measures its Euclidean distance to the nearest ground-truth object (if one exists). \textbf{NLL} is the average negative log-likelihood over the ground-truth object locations after dividing the prediction probabilities by their sum across all locations. We evaluate the metrics over 4000 crops from our HM3D val map dataset and report the results in Tab. \ref{tab:pred_ablation}. We observe that the predictions made using global context are significantly more accurate than those using single-frame inputs. Since more accurate predictions enable more efficient downstream navigation, we conclude that leveraging global context is beneficial for prediction-based navigation.

\paragraph{Distance Weighting for Goal Selection.} We study the effect of weighting the predicted probabilities by their distance to encourage selection of closer goals. 
We also evaluate an agent that does not use predictions at all and always navigates to the nearest unexplored location, which is equivalent to frontier-based exploration \cite{yamauchi1997frontier}. For this agent, we clip the minimum distance to \SI{3}{\metre} to prevent excessive back-and-forth movement. Navigation performance measured over 500 episodes from the HM3D val split is reported in Tab. \ref{tab:nav_ablation}. We find that, without distance weighting, the agent performs slightly worse because it is may ignore closer target objects in favor of farther ones. Without predictions, the agent performs significantly worse because it wastes time searching in useless locations.

\paragraph{Limitations}
A significant amount of \method{}'s failures are caused by errors in the initial semantic segmentation, as demonstrated by the large increase in performance when using ground-truth target segmentation instead of the Mask-RCNN (Tab. \ref{tab:nav_ablation}). 
Another source of failures is scenes that contain multiple floors. Since \method{} relies on a single top-down map, it is unable to traverse staircases and is prone to getting stuck within them.

\begin{table}[!t]
\caption{Effect of using global context instead of single-frame egocentric context for prediction on maps from HM3D (val).}
\small
\centering
\begin{tabular}{lcc}
\toprule
Input  & DTO ($\downarrow$) & NLL ($\downarrow$)   \\
\midrule
Egocentric Crop & 1.974 & 9.460 \\
Global Map & {\bf 1.518} & {\bf 8.873} \\
\bottomrule
\label{tab:pred_ablation}
\end{tabular}
\end{table}

\begin{table}[!t]
\caption{Ablation studies on HM3D (val). We study the effect of 1) not using distance-weighting (DW) during goal selection, 2) not using the target predictions (Pred), and 3) using ground-truth target segmentation (GT). }
\small
\centering
\begin{tabular}{ccc|cc}
\toprule
Pred  & DW  & GT Seg & SPL ($\uparrow$)  & Success ($\uparrow$) \\
\midrule
\checkmark & \checkmark & & \textbf{ 0.320 } & \textbf{0.638} \\
  \checkmark& &  & 0.310 & 0.598 \\
 & \checkmark &  & 0.235 & 0.478 \\
\bottomrule
\checkmark & \checkmark & \checkmark  & 0.400 &  0.756 \\
\bottomrule
\label{tab:nav_ablation}
\end{tabular}
\end{table}

\section{Conclusion}

We presented \method{}, a modular method for ObjectGoal navigation that predicts unseen target objects from a top-down semantic map and navigates to them. The prediction model can be trained in a supervised manner using passively collected maps. Unlike previous map prediction methods, our method takes into account the global context from previously explored areas to produce more accurate predictions, which in turn allows for more efficient navigation. Experiments showed that \method{} outperforms the state-of-the-art on standard ObjectNav benchmarks. In the future, \method{} may be improved by leveraging RGB information and or 3D geometry to build even more powerful prediction models.

{\small
\bibliographystyle{ieee_fullname}
\bibliography{egbib}
}

\end{document}

\title{Supplementary Material \\ PEANUT: Predicting and Navigating to Unseen Targets}

\maketitle

\begin{abstract}
    In the following supplementary material, we provide additional experimental details (Sec. 1) and additional qualitative results (Sec. 2 to Sec. 4).
\end{abstract}

\section{Additional Experimental Details}

We first provide additional details about our prediction network architecture, our semantic mapping module, and our procedure for finetuning Mask-RCNN on HM3D.

\subsection{Prediction Network Architecture Details}
Our target prediction network has a PSPNet-based architecture and largely follows the base implementation provided in \cite{mmseg2020}. The backbone is a ResNet50 \cite{he2016deep} with dilated convolutions as proposed in \cite{chen2017deeplab}. This outputs feature maps that are $1/8$ the size of the input. It is followed by a pyramid pooling module with pooling scales of 1, 2, 3, and 6. The fused features are passed through a $3 \times 3$ convolution layer with 512 channels and then a $1 \times 1$ convolution for final prediction. Intermediate supervision is provided through  an FCN \cite{long2015fully} prediction head before the last stage of the ResNet50.

\subsection{Semantic Mapping Details}
Our semantic mapper is essentially the same as that of SemExp \cite{chaplot2020object}. The map resolution is \SI{5}{\cm}, and the global map size is $H \times W=960 \times 960$. The 2D segmentation and depth images are downsampled by $4 \times$ before being projected to the map in order to save computation. Points within the height range $[\SI{0.25}{\metre}, \SI{0.88}{\metre}]$ are marked as obstacles. This is because \SI{0.88}{\metre} is the agent's height, and ignoring points close to the ground allows for some robustness to minor elevation changes.

\subsection{Mask-RCNN Finetuning Details}
For HM3D, we finetune a COCO-pretrained Mask-RCNN \cite{he2017mask} on images from Habitat. We collect images by letting an agent wander around according to the baseline exploration policy described in \cite{luo2022stubborn}. In total, we collect 80K images from the HM3D train split and 20K images from the HM3D val split -- 1K from each scene. We use the associated semantic annotations to extract instance segmentation masks for each image, and discard masks that are either less than 1000 pixels in area or have a bounding box whose aspect ratio is either less than 0.1 or greater than 10.

We finetune the Mask-RCNN for \num{25000} iterations using an SGD optimizer with a learning rate of 0.02, momentum of 0.9, and batch size of 16. The learning rate was decayed by a factor of 10 after 20000 iterations. During training, we apply random resizing and horizontal flipping for data augmentation.

\section{Qualitative Results on Target Prediction}
We provide additional visualizations of predictions of target objects made by \method{}'s prediction model in Fig. \ref{fig:supp_example_preds}. They demonstrate that the model uses spatial regularities and semantic cues to make highly informative predictions about unexplored areas. In many cases, the model seems to be more confident than a human would be.

\section{Qualitative Results on Navigation Episodes}
We provide additional visualizations of navigation using \method{} in Fig. \ref{fig:supp_example_episodes}.
Overall, they suggest that \method{} searches through rooms in an efficient manner. The prediction-based goal selection usually picks goals with a large amount of unexplored area nearby and does not exhibit excessive backtracking.

\section{Failure Case Analysis}

We visualize some failure cases of \method{} in Fig. \ref{fig:supp_failure_cases}. They are representative of \method{}'s most common failure modes, which are segmentation errors and scenes with stairs. In the first case, a sofa is misclassified as a chair. In the second case, a bathtub is misclassified as a bed. These result in false positive detections of the target category, causing the agent to move to the misclassified object and stop, failing the episode. In the third failure case, the target category does not exist on the agent's starting floor, so the agent must traverse a staircase to search a different floor. However, the mapping module marks stairs as obstacles, preventing their traversal.

\section{HM3D Test-Standard Leaderboard Evaluation}

We note that our method (entry name: ``Finding NIMO (PEANUT)'') currently ranks 3rd in terms of SPL on the public leaderboard for the HM3D test-standard split (as of November 16th, 2022). Since the rank 1 and 2 methods are not published, \method{}'s ranking is higher than that of any published method. The next highest ranking published method is ProcTHOR \cite{deitke2022procthor}, which relies on additional training scenes generated using the AI2-THOR simulator \cite{kolve2017ai2}. Our method is trained only on the default HM3D ObjectNav dataset, but achieves better performance.

\section{MP3D Test-Standard Leaderboard Evaluation}
We tried submitting our MP3D agent to the 2021 Habitat Challenge leaderboard on EvalAI, but our submissions repeatedly failed for unknown reasons (see screenshot in Fig. \ref{fig:mp3d_test_fail}). Thus, we report results on the MP3D val split in the main paper. We will continue to investigate the cause of this issue.

\begin{table*}[]
    \centering
    \resizebox{\linewidth}{!}{
\setlength{\tabcolsep}{0.1em} %
\renewcommand{\arraystretch}{1.}
    \begin{tabular}{cccc}
          \tikz{
        \node[draw=black, line width=.5mm, inner sep=0pt] 
            {\includegraphics[width=.22\linewidth]{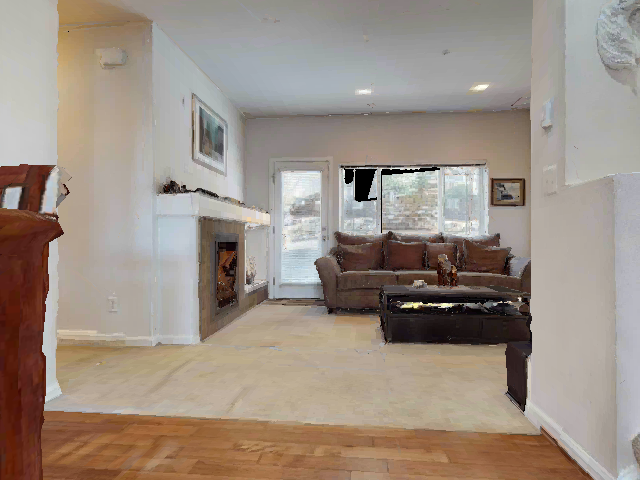}};
            \node[draw=black, draw opacity=1.0, line width=.3mm, fill opacity=0.8,fill=white, text opacity=1] at (-0.4 , 1.15) { \ Target: tv\_monitor \ };
        } &
        \tikz{
        \node[draw=black, line width=.5mm, inner sep=0pt] 
            {\includegraphics[width=.22\linewidth]{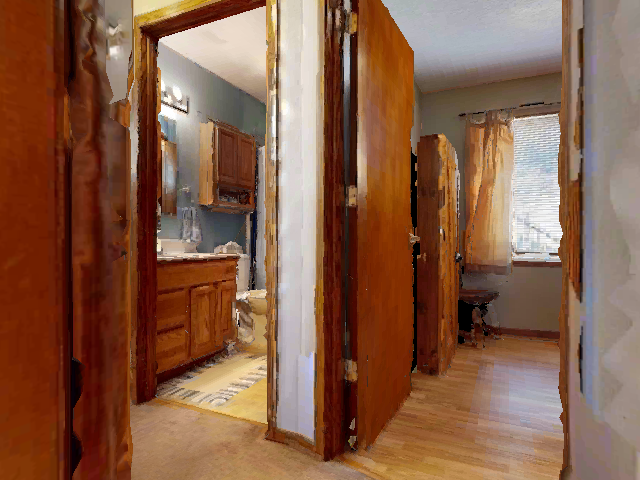}};
            \node[draw=black, draw opacity=1.0, line width=.3mm, fill opacity=0.8,fill=white, text opacity=1] at  (-0.8 , 1.15) { \ Target: chair \ };
        }& 
        \tikz{
        \node[draw=black, line width=.5mm, inner sep=0pt] 
            {\includegraphics[width=.22\linewidth]{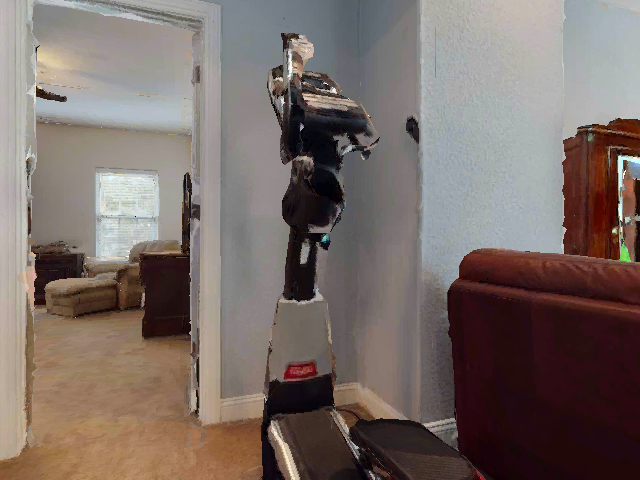}};
            \node[draw=black, draw opacity=1.0, line width=.3mm, fill opacity=0.8,fill=white, text opacity=1] at  (-0.9 , 1.15) { \ Target: bed \ };
        }& 
       \tikz{
        \node[draw=black, line width=.5mm, inner sep=0pt] 
            {\includegraphics[width=.22\linewidth]{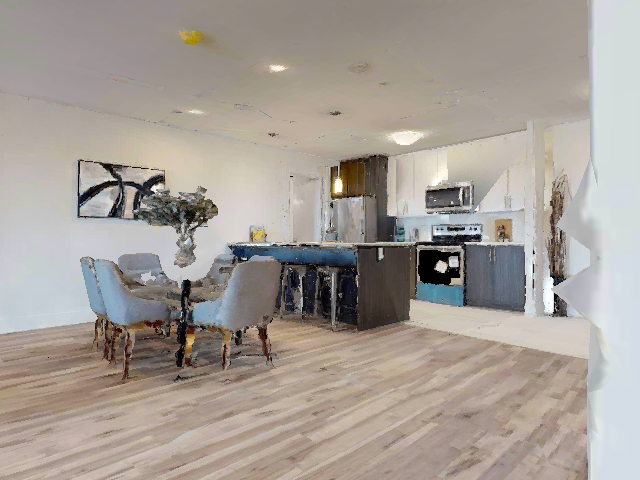}};
            \node[draw=black, draw opacity=1.0, line width=.3mm, fill opacity=0.8,fill=white, text opacity=1] at  (-0.4 , 1.15) { \ Target: tv\_monitor \ };
        } 
        \\
        
        \tikz{
        \node[draw=black, line width=.5mm, inner sep=0pt] 
        {\includegraphics[trim=0.75cm 0.75cm 0.75cm 0.75cm, clip,width=.22\linewidth]{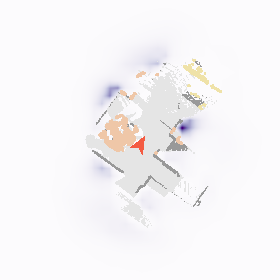}};
        } &
        \tikz{
        \node[draw=black, line width=.5mm, inner sep=0pt] 
        {\includegraphics[trim=0.75cm 0.75cm 0.75cm 0.75cm, clip,width=.22\linewidth]{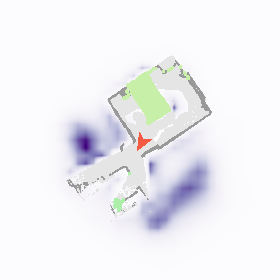}};
        }  & 
        \tikz{
        \node[draw=black, line width=.5mm, inner sep=0pt] 
        {\includegraphics[trim=0.75cm 0.75cm 0.75cm 0.75cm, clip,width=.22\linewidth]{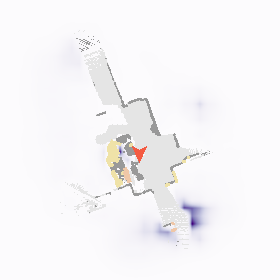}};
        }  & 
        \tikz{
        \node[draw=black, line width=.5mm, inner sep=0pt] 
        {\includegraphics[trim=0.75cm 0.75cm 0.75cm 0.75cm, clip,width=.22\linewidth]{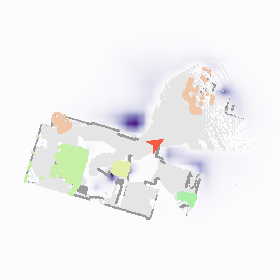}};
        } \\
        
          \tikz{
        \node[draw=black, line width=.5mm, inner sep=0pt] 
            {\includegraphics[width=.22\linewidth]{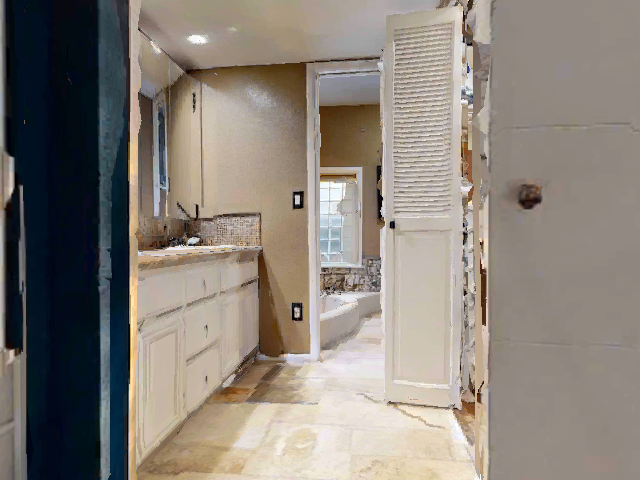}};
            \node[draw=black, draw opacity=1.0, line width=.3mm, fill opacity=0.8,fill=white, text opacity=1] at (-0.8 , 1.15) { \ Target: toilet \ };
        } &
        \tikz{
        \node[draw=black, line width=.5mm, inner sep=0pt] 
            {\includegraphics[width=.22\linewidth]{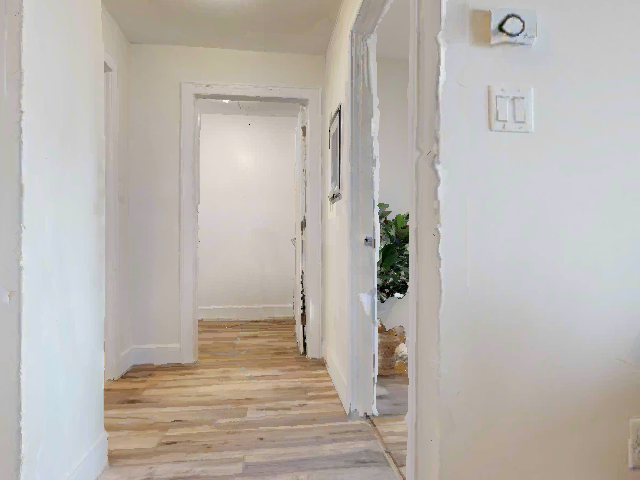}};
            \node[draw=black, draw opacity=1.0, line width=.3mm, fill opacity=0.8,fill=white, text opacity=1] at  (-0.9 , 1.15) { \ Target: bed \ };
        }& 
        \tikz{
        \node[draw=black, line width=.5mm, inner sep=0pt] 
            {\includegraphics[width=.22\linewidth]{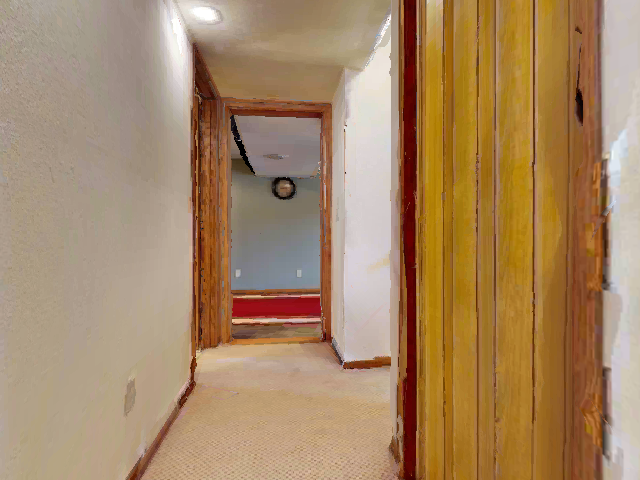}};
            \node[draw=black, draw opacity=1.0, line width=.3mm, fill opacity=0.8,fill=white, text opacity=1] at  (-0.8 , 1.15) { \ Target: chair \ };
        }& 
       \tikz{
        \node[draw=black, line width=.5mm, inner sep=0pt] 
            {\includegraphics[width=.22\linewidth]{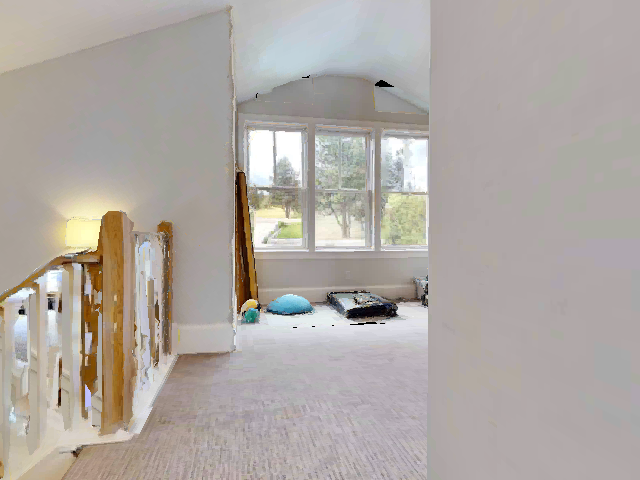}};
            \node[draw=black, draw opacity=1.0, line width=.3mm, fill opacity=0.8,fill=white, text opacity=1] at  (-0.85 , 1.15) { \ Target: sofa \ };
        } 
        \\
        
        \tikz{
        \node[draw=black, line width=.5mm, inner sep=0pt] 
        {\includegraphics[trim=0.75cm 0.75cm 0.75cm 0.75cm, clip,width=.22\linewidth]{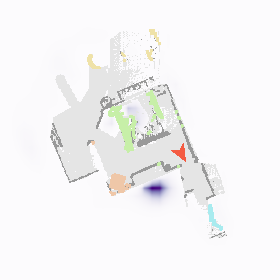}};
        } &
        \tikz{
        \node[draw=black, line width=.5mm, inner sep=0pt] 
        {\includegraphics[trim=0.75cm 0.75cm 0.75cm 0.75cm, clip,width=.22\linewidth]{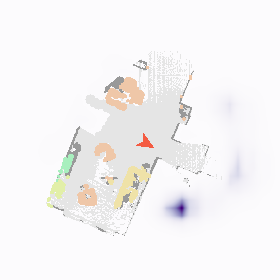}};
        }  & 
        \tikz{
        \node[draw=black, line width=.5mm, inner sep=0pt] 
        {\includegraphics[trim=0.75cm 0.75cm 0.75cm 0.75cm, clip,width=.22\linewidth]{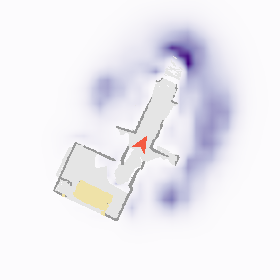}};
        }  & 
        \tikz{
        \node[draw=black, line width=.5mm, inner sep=0pt] 
        {\includegraphics[trim=0.75cm 0.75cm 0.75cm 0.75cm, clip,width=.22\linewidth]{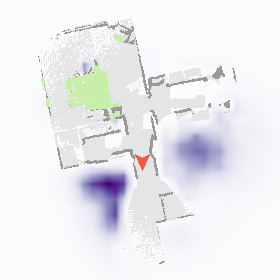}};
        } \\
        
    \end{tabular}
    }
    \includegraphics[width=0.9\linewidth]{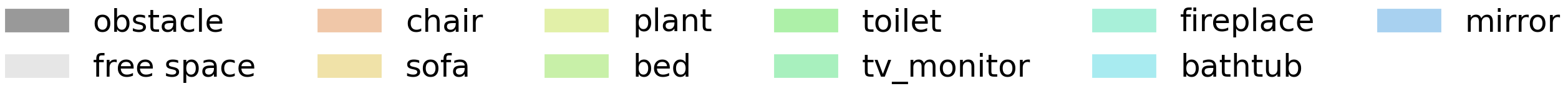}
\captionof{figure}{\textbf{Example target predictions}. We visualize predictions made by our model in scenes from HM3D (val). The top row shows the agent's RGB observation, and the bottom row shows the incomplete semantic map overlaid with the target probability prediction. Note that the prediction network only has access to the semantic map, not the RGB images.
}
\label{fig:supp_example_preds}
\end{table*}

\begin{table*}[]
    \centering
    \resizebox{\linewidth}{!}{
\setlength{\tabcolsep}{0.1em} %

    \begin{tabular}{ccccc}
        \tikz{
        \node[draw=black, line width=.5mm, inner sep=0pt] 
            {\includegraphics[width=.22\linewidth]{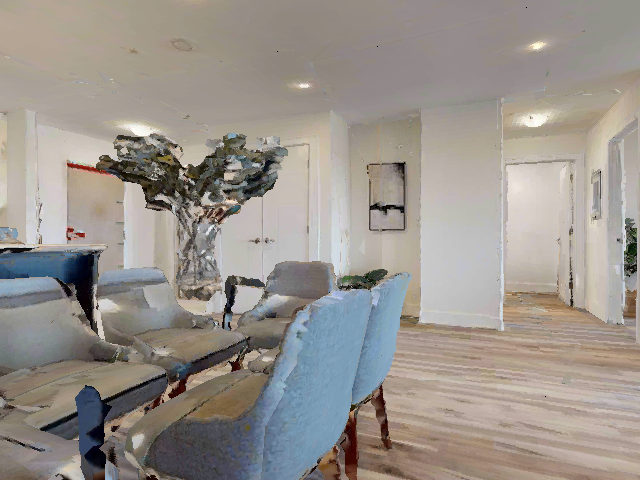}};
            \node[draw=black, draw opacity=1.0, line width=.3mm, fill opacity=0.8,fill=white, text opacity=1] at (-0.90 , 1.15) { \ Target: bed \ };
        } &
        \tikz{
        \node[draw=black, line width=.5mm, inner sep=0pt] 
        {\includegraphics[width=.22\linewidth]{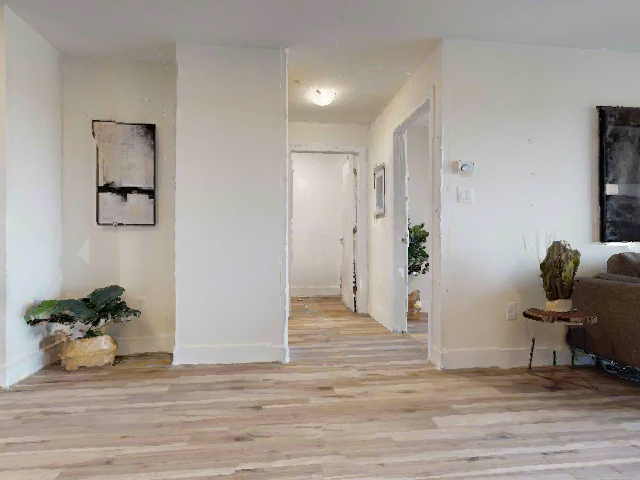}};
        } & 
        \tikz{
        \node[draw=black, line width=.5mm, inner sep=0pt] 
        {\includegraphics[width=.22\linewidth]{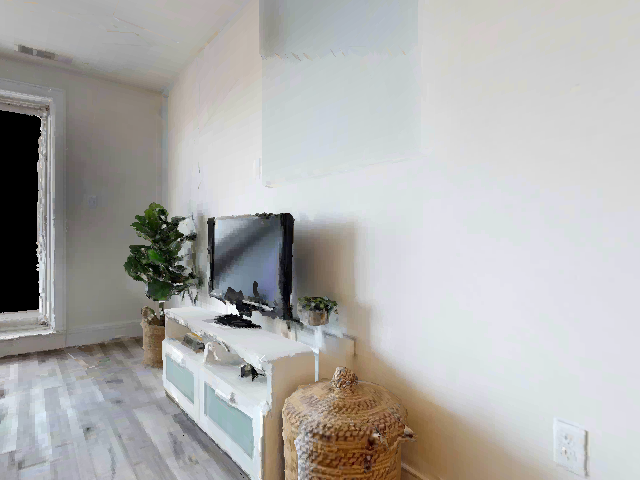}};
        } & 
        \tikz{
        \node[draw=black, line width=.5mm, inner sep=0pt] 
        {\includegraphics[width=.22\linewidth]{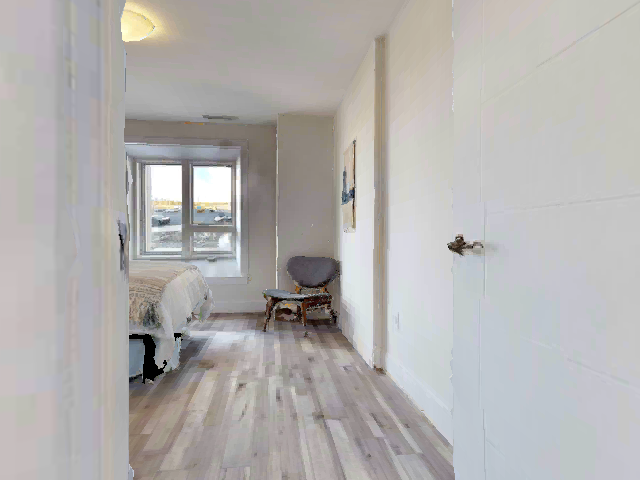}};
        } & 
        \tikz{
        \node[draw=black, line width=.5mm, inner sep=0pt] 
        {\includegraphics[width=.22\linewidth]{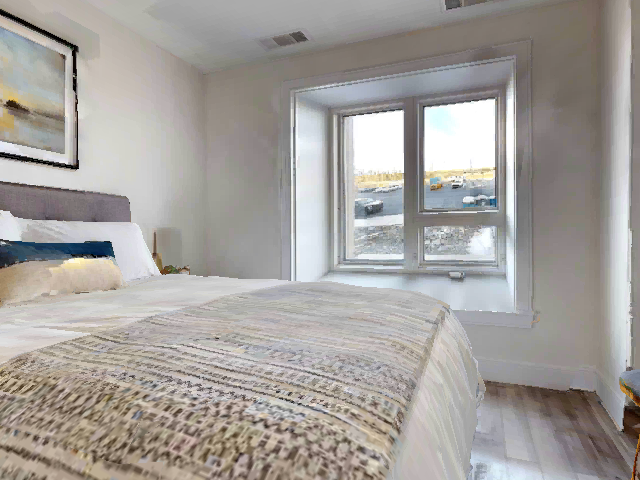}};
        } \\
        
        \tikz{
        \node[draw=black, line width=.5mm, inner sep=0pt] 
        {\includegraphics[trim=0cm 0cm 0cm 0cm, clip,width=.22\linewidth]{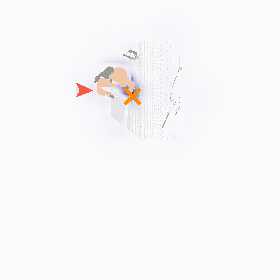}};
        } &
        \tikz{
        \node[draw=black, line width=.5mm, inner sep=0pt] 
        {\includegraphics[trim=0cm 0cm 0cm 0cm, clip,width=.22\linewidth]{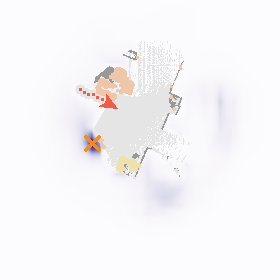}};
        } & 
        \tikz{
        \node[draw=black, line width=.5mm, inner sep=0pt] 
        {\includegraphics[trim=0cm 0cm 0cm 0cm, clip,width=.22\linewidth]{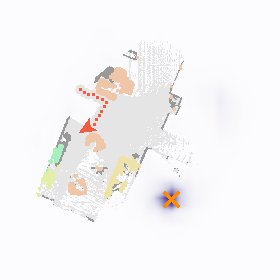}};
        } & 
        \tikz{
        \node[draw=black, line width=.5mm, inner sep=0pt] 
        {\includegraphics[trim=0cm 0cm 0cm 0cm, clip,width=.22\linewidth]{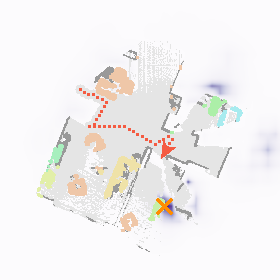}};
        } & 
        \tikz{
        \node[draw=black, line width=.5mm, inner sep=0pt] 
        {\includegraphics[trim=0cm 0cm 0cm 0cm, clip,width=.22\linewidth]{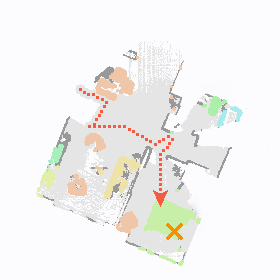}};
        } \\
        $t=1$ & 
        $t=10$ & 
        $t=20$ & 
        $t=65$ & 
        $t=78$  \\
        
        \midrule
        \tikz{
        \node[draw=black, line width=.5mm, inner sep=0pt] 
            {\includegraphics[width=.22\linewidth]{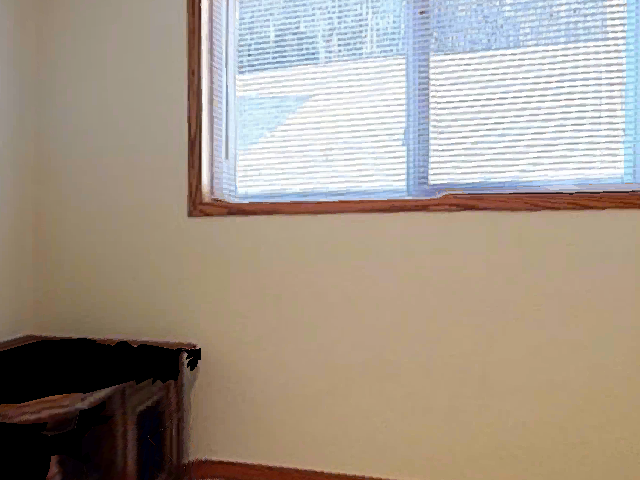}};
            \node[draw=black, draw opacity=1.0, line width=.3mm, fill opacity=0.8,fill=white, text opacity=1] at (-0.80 , 1.15) { \ Target: chair \ };
        } &
        \tikz{
        \node[draw=black, line width=.5mm, inner sep=0pt] 
        {\includegraphics[width=.22\linewidth]{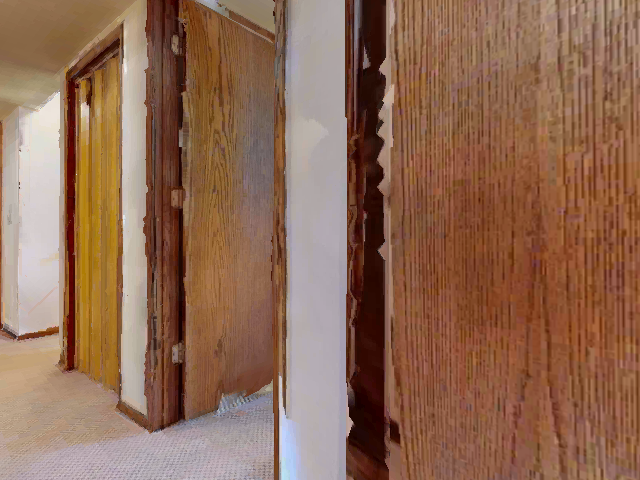}};
        } & 
        \tikz{
        \node[draw=black, line width=.5mm, inner sep=0pt] 
        {\includegraphics[width=.22\linewidth]{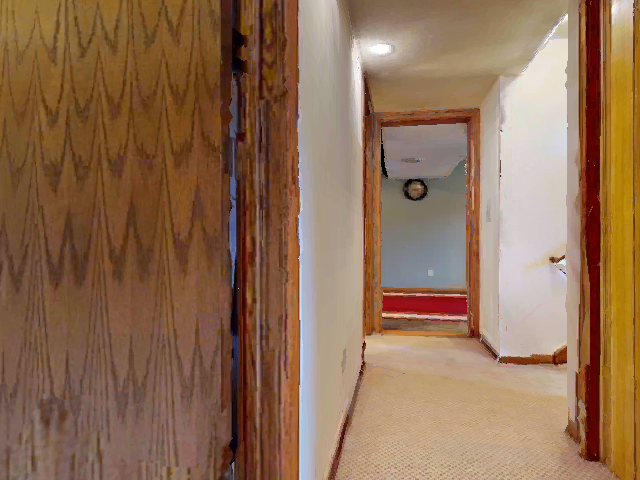}};
        } & 
        \tikz{
        \node[draw=black, line width=.5mm, inner sep=0pt] 
        {\includegraphics[width=.22\linewidth]{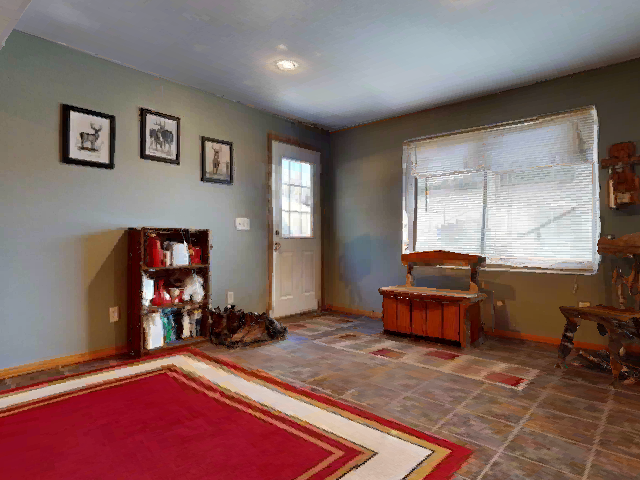}};
        } & 
        \tikz{
        \node[draw=black, line width=.5mm, inner sep=0pt] 
        {\includegraphics[width=.22\linewidth]{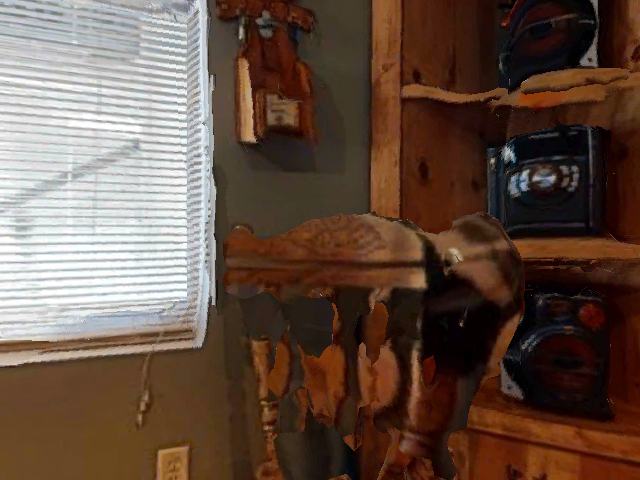}};
        } \\
        
        \tikz{
        \node[draw=black, line width=.5mm, inner sep=0pt] 
        {\includegraphics[trim=0cm 0cm 0cm 0cm, clip,width=.22\linewidth]{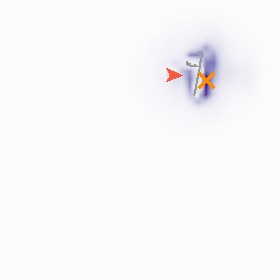}};
        } &
        \tikz{
        \node[draw=black, line width=.5mm, inner sep=0pt] 
        {\includegraphics[trim=0cm 0cm 0cm 0cm, clip,width=.22\linewidth]{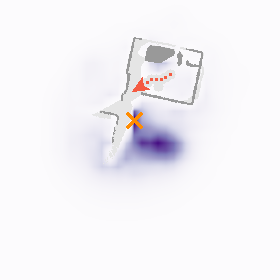}};
        } & 
        \tikz{
        \node[draw=black, line width=.5mm, inner sep=0pt] 
        {\includegraphics[trim=0cm 0cm 0cm 0cm, clip,width=.22\linewidth]{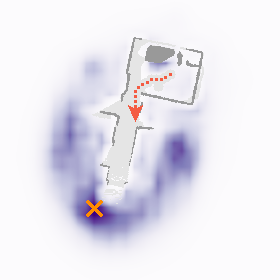}};
        } & 
        \tikz{
        \node[draw=black, line width=.5mm, inner sep=0pt] 
        {\includegraphics[trim=0cm 0cm 0cm 0cm, clip,width=.22\linewidth]{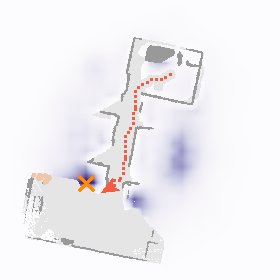}};
        } & 
        \tikz{
        \node[draw=black, line width=.5mm, inner sep=0pt] 
        {\includegraphics[trim=0cm 0cm 0cm 0cm, clip,width=.22\linewidth]{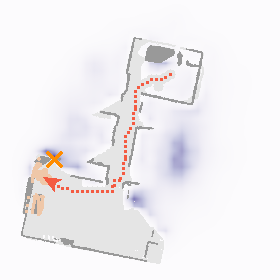}};
        } \\
        $t=1$ & 
        $t=30$ & 
        $t=40$ & 
        $t=85$ & 
        $t=118$  \\
        
        \midrule
        \tikz{
        \node[draw=black, line width=.5mm, inner sep=0pt] 
            {\includegraphics[width=.22\linewidth]{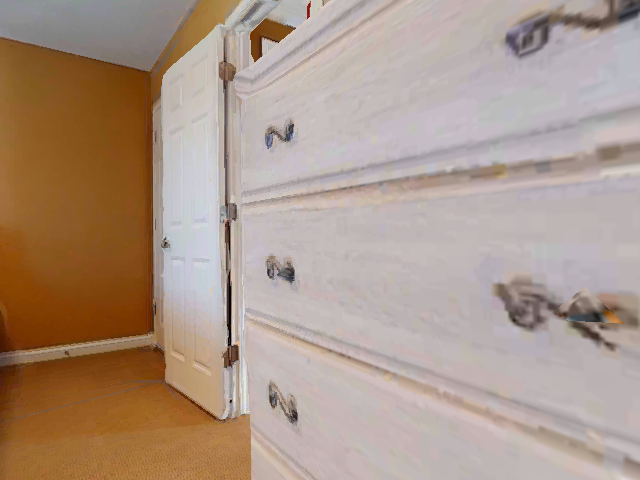}};
            \node[draw=black, draw opacity=1.0, line width=.3mm, fill opacity=0.8,fill=white, text opacity=1] at (-0.40 , 1.15) { \ Target: tv_monitor \ };
        } &
        \tikz{
        \node[draw=black, line width=.5mm, inner sep=0pt] 
        {\includegraphics[width=.22\linewidth]{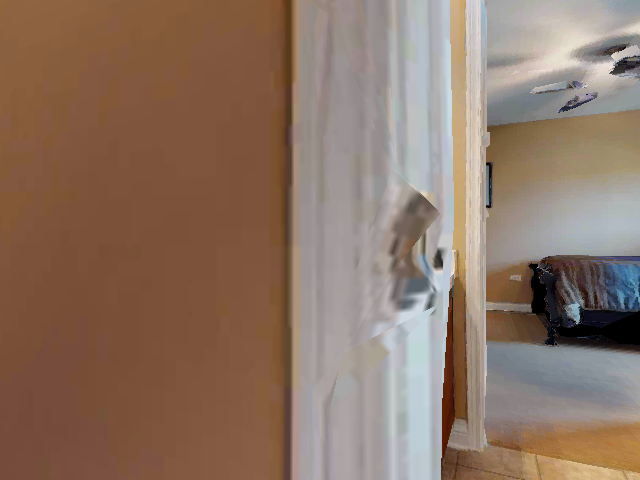}};
        } & 
        \tikz{
        \node[draw=black, line width=.5mm, inner sep=0pt] 
        {\includegraphics[width=.22\linewidth]{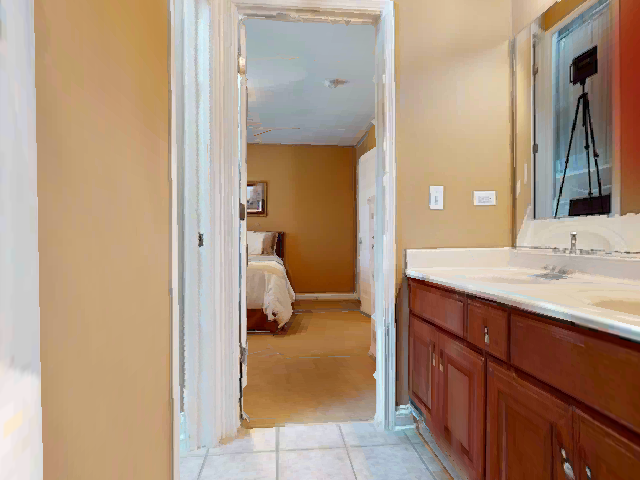}};
        } & 
        \tikz{
        \node[draw=black, line width=.5mm, inner sep=0pt] 
        {\includegraphics[width=.22\linewidth]{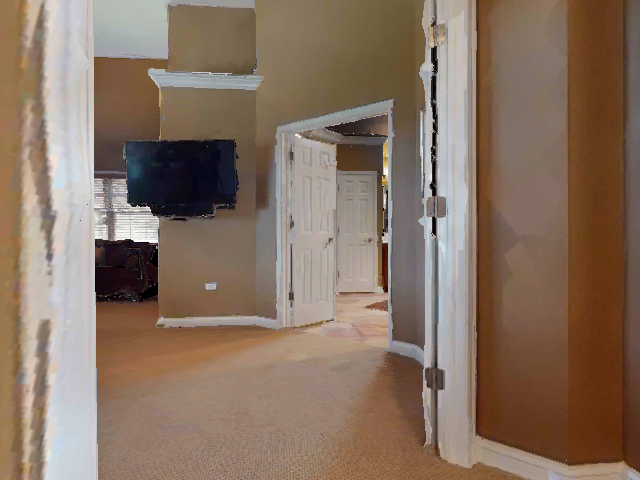}};
        } & 
        \tikz{
        \node[draw=black, line width=.5mm, inner sep=0pt] 
        {\includegraphics[width=.22\linewidth]{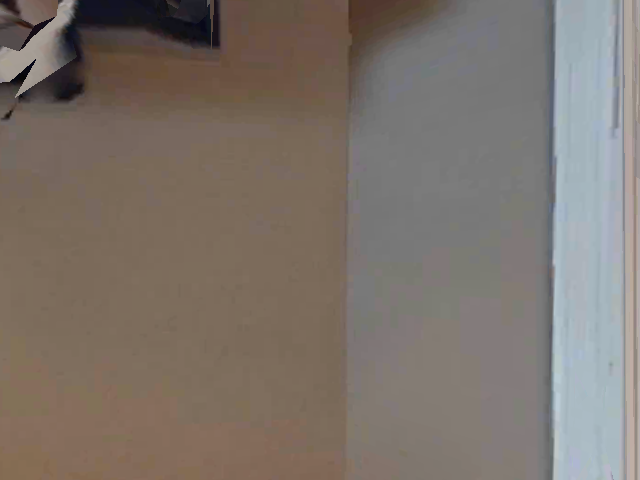}};
        } \\
        
        \tikz{
        \node[draw=black, line width=.5mm, inner sep=0pt] 
        {\includegraphics[trim=0cm 0cm 0cm 0cm, clip,width=.22\linewidth]{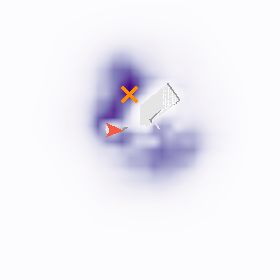}};
        } &
        \tikz{
        \node[draw=black, line width=.5mm, inner sep=0pt] 
        {\includegraphics[trim=0cm 0cm 0cm 0cm, clip,width=.22\linewidth]{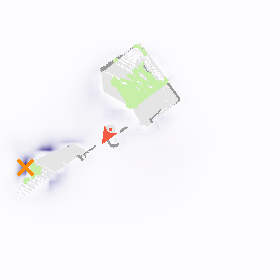}};
        } & 
        \tikz{
        \node[draw=black, line width=.5mm, inner sep=0pt] 
        {\includegraphics[trim=0cm 0cm 0cm 0cm, clip,width=.22\linewidth]{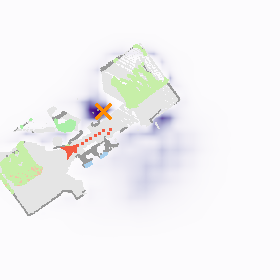}};
        } & 
        \tikz{
        \node[draw=black, line width=.5mm, inner sep=0pt] 
        {\includegraphics[trim=0cm 0cm 0cm 0cm, clip,width=.22\linewidth]{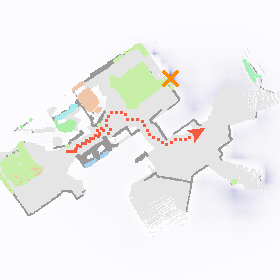}};
        } & 
        \tikz{
        \node[draw=black, line width=.5mm, inner sep=0pt] 
        {\includegraphics[trim=0cm 0cm 0cm 0cm, clip,width=.22\linewidth]{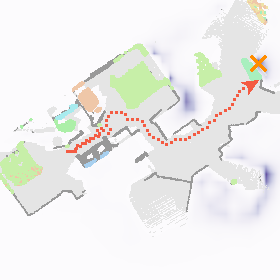}};
        } \\
        $t=1$ & 
        $t=10$ & 
        $t=50$ & 
        $t=126$ & 
        $t=147$  \\
    \end{tabular}
    }
\captionof{figure}{\textbf{Example navigation episodes.} We visualize episodes of navigation in scenes from HM3D (val). The top row shows the agent's RGB observation, and the bottom row shows the incomplete semantic map overlaid with the target probability prediction. The agent's selected long-term goal is marked by an orange cross. 
}
\label{fig:supp_example_episodes}
\end{table*}

\begin{table*}[]
    \centering
    \resizebox{\linewidth}{!}{
\setlength{\tabcolsep}{0.1em} %
\renewcommand{\arraystretch}{1.}
    \begin{tabular}{ccc}
          \tikz{
        \node[draw=black, line width=.5mm, inner sep=0pt] 
            {\includegraphics[width=.22\linewidth]{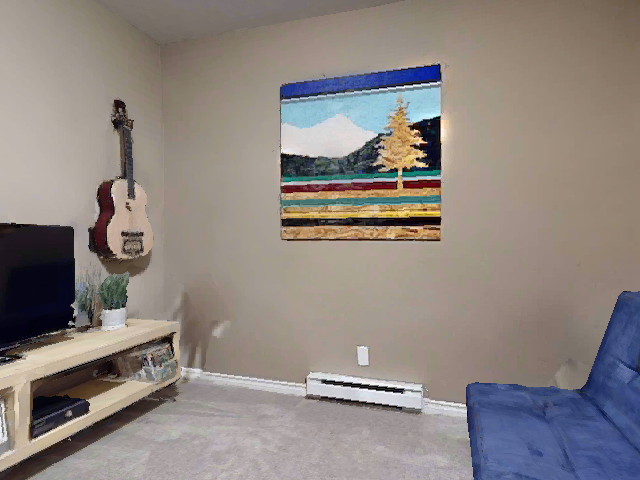}};
            \node[draw=black, draw opacity=1.0, line width=.3mm, fill opacity=0.8,fill=white, text opacity=1] at (-0.82 , 1.15) { \ Target: chair \ };
        } &
        \tikz{
        \node[draw=black, line width=.5mm, inner sep=0pt] 
            {\includegraphics[width=.22\linewidth]{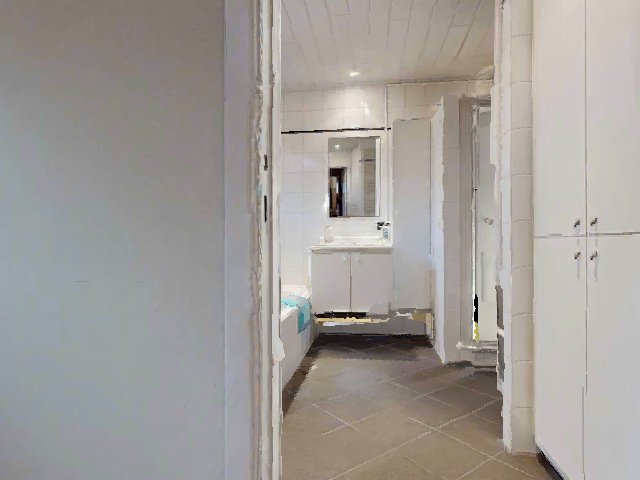}};
            \node[draw=black, draw opacity=1.0, line width=.3mm, fill opacity=0.8,fill=white, text opacity=1] at  (-0.9 , 1.15) { \ Target: bed \ };
        }& 
        \tikz{
        \node[draw=black, line width=.5mm, inner sep=0pt] 
            {\includegraphics[width=.22\linewidth]{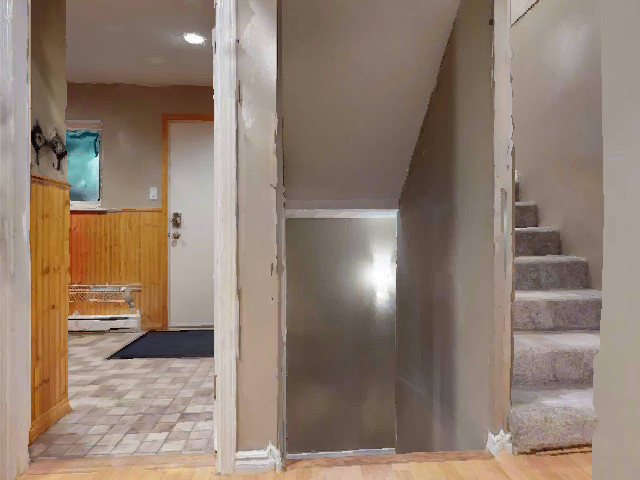}};
            \node[draw=black, draw opacity=1.0, line width=.3mm, fill opacity=0.8,fill=white, text opacity=1] at  (-0.9 , 1.15) { \ Target: bed \ };
        }
        \\
        
        \tikz{
        \node[draw=black, line width=.5mm, inner sep=0pt] 
        {\includegraphics[trim=0.75cm 0.75cm 0.75cm 0.75cm, clip,width=.22\linewidth]{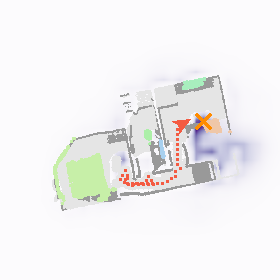}};
        } &
        \tikz{
        \node[draw=black, line width=.5mm, inner sep=0pt] 
        {\includegraphics[trim=0.75cm 0.75cm 0.75cm 0.75cm, clip,width=.22\linewidth]{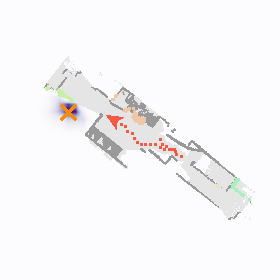}};
        }  & 
        \tikz{
        \node[draw=black, line width=.5mm, inner sep=0pt] 
        {\includegraphics[trim=0.75cm 0.75cm 0.75cm 0.75cm, clip,width=.22\linewidth]{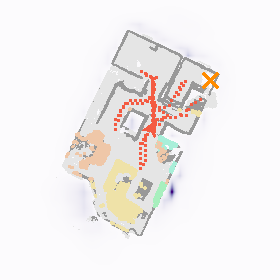}};
        }  \\
        
    \end{tabular}
    }
\captionof{figure}{\textbf{Example failure cases}. We visualize some of \method{}'s failure cases on HM3D (val). The first two are caused by false positive detections of the target. The third results from the agent's inability to traverse stairs and search in another floor (due to the stairs appearing as obstacles on the agent's map).
}
\label{fig:supp_failure_cases}

\vspace{60pt}
\includegraphics[width=0.85\linewidth]{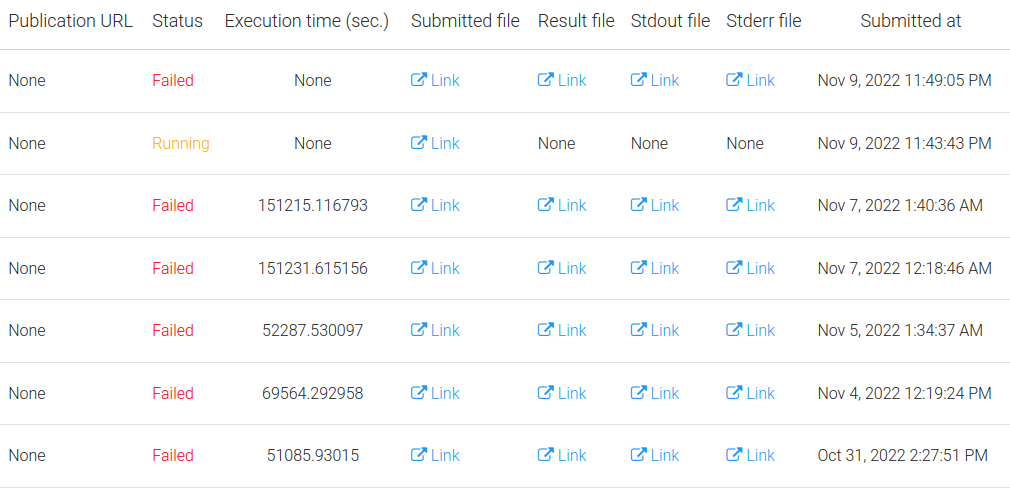}
\caption{Screenshot of our submission history to the MP3D test-standard split leaderboard as of November 16th, 2022.}
\label{fig:mp3d_test_fail}
\end{table*}

{\small
\bibliographystyle{ieee_fullname}
\bibliography{egbib}
}